%% file: iclr2027_conference.tex
\documentclass{article} % For LaTeX2e
\usepackage{arxiv,times}

\input{math_commands.tex}

\usepackage{hyperref}
\usepackage{url}
\usepackage{xspace}
\usepackage{booktabs}
\usepackage[table]{xcolor}
\usepackage{subcaption}
\usepackage{fontawesome5}
\usepackage{enumitem}
\usepackage{wrapfig}
\usepackage{graphicx} 
\usepackage{marvosym}

\newcommand{\name}{\textbf{Robo\textcolor{orange}{Dawn}}}

\title{
\raisebox{-0.2\height}{\includegraphics[width=0.09\textwidth]{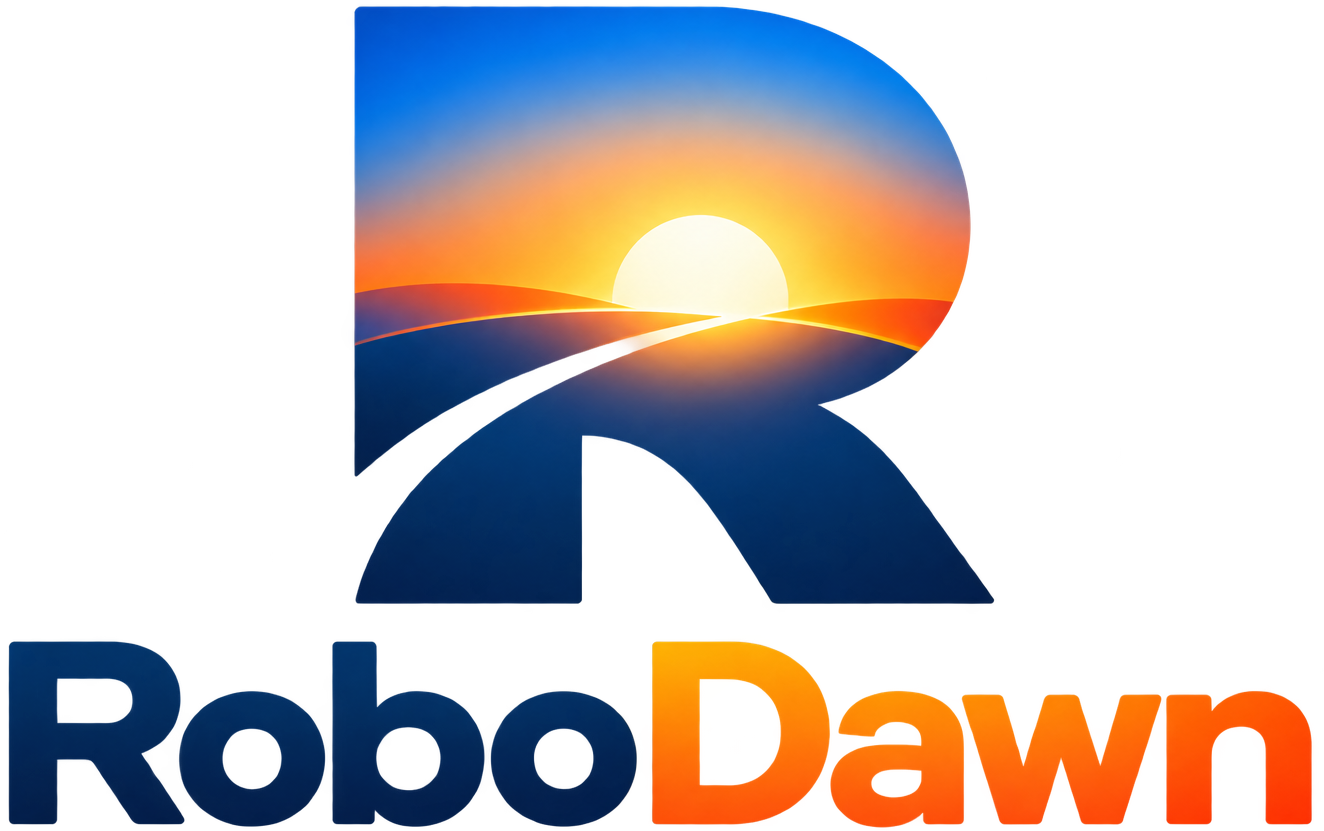}}
Transferring the Intelligence of VLMs to Robotic Control}

\newcommand{\projecturl}{%
  \href{https://robodawn.top/}{%
    \faHome\ 
    \texttt{https://}
    \texttt{Robo\textcolor{orange}{Dawn}}
    \texttt{.top/}%
  }%
}

\author{%
  \begin{minipage}[t]{\dimexpr\textwidth-2\tabcolsep\relax}
    \centering
    \normalfont\normalsize
    \textbf{Meng-Hao Guo}$^{1}$\quad
    \textbf{Zhe-Han Mo}$^{1}$\quad
    \textbf{Jia-Jun Wang}$^{1}$\quad
    \textbf{Yi Zhang}$^{1}$\quad
    \textbf{Kejin Wang}$^{1}$
    \\[3pt]
    \textbf{Yi-Xuan Deng}$^{1}$\quad
    \textbf{Jia-Peng Zhang}$^{1}$\quad
    \textbf{Yongming Rao}$^{2}$\quad
    \textbf{Shi-Min Hu}$^{1}$\thanks{Corresponding author.}
    \\[6pt]
    $^{1}$Tsinghua University
    \qquad
    $^{2}$Tencent Hunyuan
    \\[4pt]
    \makebox[\textwidth][c]{\projecturl}
  \end{minipage}%
}

\iclrfinalcopy % Uncomment for camera-ready version, but NOT for submission.
\begin{document}

\maketitle

\begin{abstract}
Humans can seamlessly adapt to both physical and digital worlds, suggesting that while a digital-to-real gap exists in embodiment, environment and task, human intelligence itself may transfer across this gap.
% This naturally raises a fundamental question: can the intelligence of vision-language models (VLMs) similarly generalize from the digital world to the physical world 
% for robotics control?
This naturally raises a fundamental question: can the intelligence of vision-language models (VLMs) similarly generalize from the digital world to the physical world for robotic control?
We investigate this question through \textbf{\name}, a human-intuitive interface that exposes robotic control to an agentic VLM through a compact set of discrete translation, rotation, and gripper commands.
Using this interface, the VLM controls a robot in a closed loop: it observes the current visual state, reasons about the next action, executes it, and adapts subsequent decisions to the resulting state.
% In this work, we propose \textbf{\name}, a human-intuitive interface that exposes robotic control to an agentic VLM through discrete translational, rotational, and gripper commands.
% The VLM then operates the robot in a closed loop, repeatedly observing the current visual state, reasoning about what to do next, executing it, and adapting its subsequent decisions based on the resulting state.
Furthermore, we introduce an in-context learning (ICL) scheme that uses a few demonstrations to ground the VLM in both interface usage and task-solving strategies.
% without task-specific training.
% Without task-specific training, VLMs can zero-shot control robots across diverse tasks, environments, and embodiments, achieving XX.X\% on RoboTwin and YY.Y\% on RoboDojo. These results substantially outperform existing zero-shot agentic VLAs approaches, while even approaching the performance of state-of-the-art(SOTA) fine-tuned models. 
% 
Experiments on RoboTwin 2.0 C2R and RoboDojo demonstrate that \name{} achieves strong performance without task-specific robot training.
In the zero-shot setting, \name{} outperforms several strong policies trained on benchmark-specific robot data, while a single in-context demonstration further yields substantial performance gains and establishes state-of-the-art (SOTA) results.
% On RoboTwin 2.0 C2R, for example, the success rate improves from 53.2\% zero-shot to 73.6\% one-shot, outperforming the previous best method $\pi_{0.5}$ by 27.6\%.
% Similar gains are observed on RoboDojo, where success rate improves from 25.40\% zero-shot to XX.X\% one-shot.
On RoboTwin 2.0 C2R, the success rate increases from 53.2\% zero-shot to 73.6\% one-shot, exceeding the solid baseline $\pi_{0.5}$ (46.0\%).
Similar gains are observed on RoboDojo, where success rate improves from 35.67\% zero-shot to 47.17\% one-shot.
The same framework also transfers to real-world robots, performing block-in-basket and block stacking on Franka.
% Experiments demonstrate that \name{} achieves one-shot success rates of 73.6\% on RoboTwin2.0 C2R and XX.X\% on RoboDojo, respectively, while significantly outperforming strong robot policies trained on robot data, including $\pi_{0.5}$ and Lingbot-VLA.
% The same approach transfers to real-world manipulation, achieving XX.X\% success on cloth folding and YY.Y\% on pick-and-place. These results suggest that, given an accessible action interface and a few in-context examples, pretrained multimodal intelligence can be effectively translated into embodied action, offering a complementary path toward general-purpose robotic manipulation.
% Beyond simulation, our method also demonstrates strong real-world performance, achieving success rates of XX.X\% and YY.Y\% on real-robot cloth-folding and pick-and-place tasks, respectively.
% Our results suggest that, with an accessible action interface and few-shot samples, the \textbf{intelligence} acquired by VLMs can transfer directly to embodied control, positioning intelligence transfer as a promising route toward general-purpose robotics.
% Code will be available at \url{https://robodawn.top/}.
\end{abstract}

% Man is a Tool-using Animal; without tools he is nothing, with tools he is all

% \begin{quote}
% \centering
% \textit{``Man is a tool-using animal. Without tools he is nothing, with tools he is all.''}

% \begin{flushright}
% \textit{-- Thomas Carlyle}
% \end{flushright}
% \end{quote}

\section{Introduction}
\label{intro}

Humans can routinely transfer knowledge and skills between digital and physical worlds. Despite substantial differences in embodiment, environment, and task, the underlying capabilities for perception, learning, reasoning, and decision-making often remain reusable. 
For example, a person can readily adapt to controlling unfamiliar virtual players in games or robotic embodiments in the physical world across diverse environments and tasks.
We refer to this ability to reuse such general capabilities across variations in embodiment, environment, and task as \textbf{intelligence transfer}. This raises a natural question: can VLMs, whose intelligence has emerged predominantly from the digital world, exhibit a similar form of transfer and directly control robots in the physical world?

\begin{figure}[t]
\centering
\includegraphics[width=\textwidth]{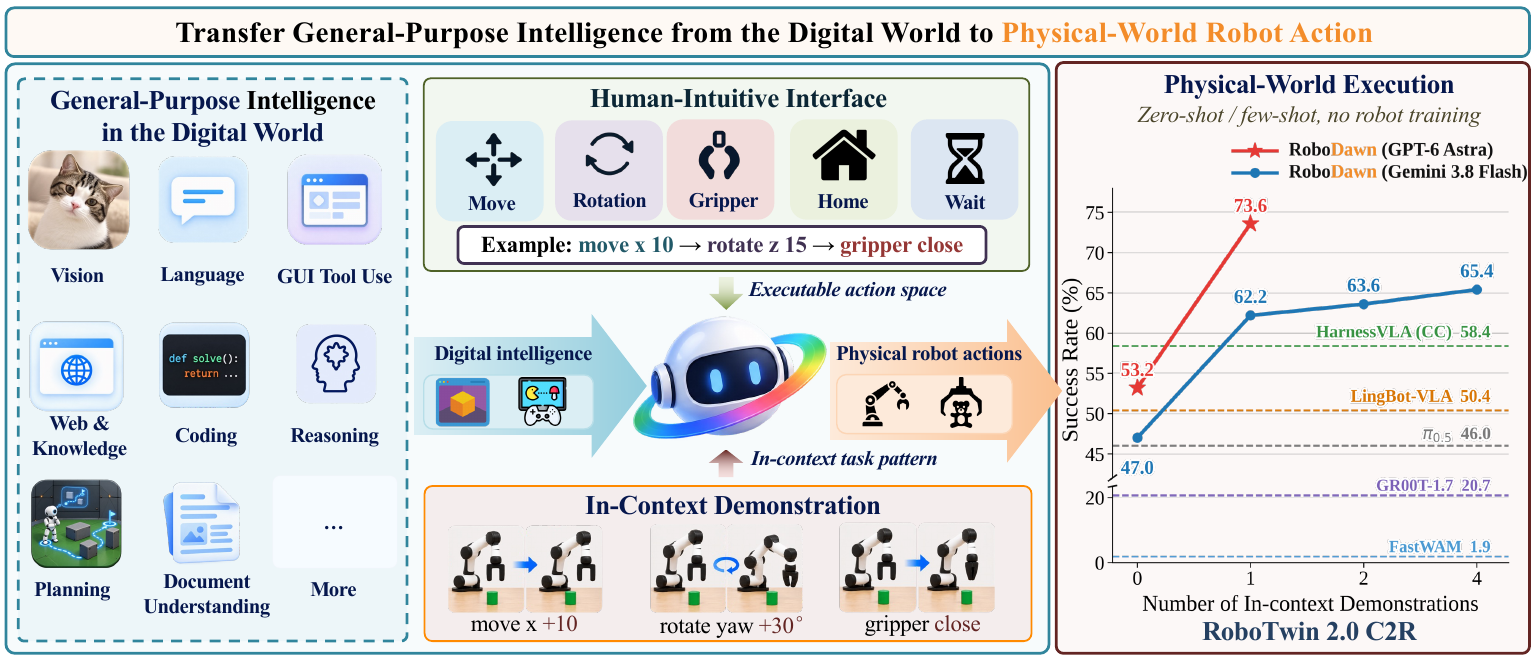}
\caption{\name{} transfers general-purpose VLM intelligence from the digital world to physical robot control through a human-intuitive action interface and in-context demonstrations.}
\label{fig:teaser}
\end{figure}

% Most existing approaches to general-purpose robotic manipulation rely on explicitly training models with robot data. 
Some existing approaches pursue this goal by adapting pretrained models such as VLMs to robotics through explicit training on robot data.
Vision-language-action (VLA) models~\citep{DBLP:conf/corl/KimPKXB0RFSVKBT24,pmlr-v229-zitkovich23a} and world-action models (WAMs)~\citep{DBLP:journals/corr/abs-2602-15922,yuan2026fast} typically learn mappings from visual and linguistic observations to robot actions using large-scale collections of paired data, which are costly to collect, embodiment-specific, and difficult to scale across robots, environments, and tasks.
More importantly, recent evidence suggests a deeper limitation: training VLMs for action prediction may lead to degradation in general-purpose capabilities.
% acquired during training,
including instruction following and reasoning~\citep{hancock2026actions,yang2026vision}, 
% while biasing the model toward the action distributions during robot training. This is particularly concerning because robot interaction data remain orders of magnitude scarcer than the language-vision data used to build modern VLMs.
This is particularly concerning because robot data remain orders of magnitude scarcer than the language-vision data used to build modern VLMs, raising the risk that large-scale parameter updates overfit to narrow robot data distributions and compromise the model’s broadly generalizable intelligence.
% Rather than forcing VLMs to relearn behavior from a comparatively narrow action corpus, it is therefore natural to ask whether their existing intelligence can be transferred directly to robotic control.
As shown in Fig.~\ref{fig:teaser}, these observations motivate a question: rather than adapting a VLM with large-scale robot training data, can we achieve 
\textbf{intelligence transfer} with a lightweight interface and only a few in-context demonstrations?

In this work, we investigate this hypothesis with \name, a human-intuitive interface that exposes robotic control to an agentic VLM through a compact set of discrete motion primitives. 
The interface provides translation, rotation, and gripper actions, each corresponding to a discrete incremental change in robot state, such as moving the end effector by multiples of a unit distance along a specified direction or rotating it by multiples of a unit angle.
Instead of predicting low-level continuous controls directly, the VLM interacts with the robot through these semantically interpretable actions. It repeatedly observes the current visual state, reasons about the next action, executes that action, and observes its consequence before deciding what to do next. Robotic manipulation is thus formulated as a closed-loop visual decision-making process that is structurally similar to interactive environments in which modern multimodal models already exhibit strong reasoning capabilities.

A simple interface along with language rules, however, does not eliminate all ambiguity~\cite{yu2022using}.
The model must understand the semantics and granularity of individual actions, infer how actions alter the observed scene, and learn interaction conventions that may not be obvious from the action names themselves. Much like humans, who may struggle to fully understand certain rules from written instructions alone but can grasp them much more easily after observing a few demonstrations, VLMs may also benefit substantially from example-based guidance.
We therefore complement \name{} with an ICL scheme in which a small number of demonstrations serve as examples of how to interact through the interface. These demonstrations require no parameter updates; instead, they provide task and interface context that helps the VLM infer effective action sequences. 
Success under these conditions would suggest that the model need not acquire embodied intelligence entirely from scratch; part of the required intelligence may already be present and only needs an appropriate mechanism to unlock it.

We evaluate \name{} across diverse simulated and real-world manipulation settings spanning tasks, environments, and robot embodiments. In simulation, we study both RoboTwin 2.0 C2R and RoboDojo under zero-shot and one-shot settings. 
% Despite receiving no task-specific robot training, \name{} achieves strong performance against methods trained or post-trained on benchmark-specific robot demonstrations. 
On RoboTwin 2.0 C2R,
% as shown in Fig.~\ref{fig:teaser}, 
\name{} achieves 53.2\% success zero-shot, already outperforming strong robot-trained policies such as $\pi_{0.5}$ (46.0\%) and LingBot-VLA (50.4\%), and improves substantially to 73.6\% with only one in-context demonstration. 
Consistent improvements are observed on 
%  the 19-task subset of 
RoboDojo, with the success rate rising from 35.67\% zero-shot to 47.17\% one-shot.
We additionally deploy the same framework on real robots, performing block-in-basket and block stacking with a Franka robot.
% , and cloth folding with dual Piper arms, providing further evidence of transfer across tasks and embodiments.
Together, these results suggest that pretrained multimodal intelligence can be translated into physical action through an appropriate interface and lightweight in-context adaptation.
% , without conventional task-specific robot-policy training.

Our results suggest a different perspective on the role of foundation models in robotics.
Much of the current effort in generalist robotics focuses on expanding robot datasets and training specific action models.
These directions are important, especially for low-level dexterity and high-frequency control.
Our findings are complementary: for a broad class of manipulation problems, a strong pretrained VLM can already serve as the high-level decision-making engine, while a simple action interface and a few demonstrations can transfer its intelligence to robotic control.
From this perspective, the key bottleneck may not always be acquiring new embodied intelligence from scratch, but rather designing effective interfaces and online lessons that enable existing intelligence to transfer, act, and improve in the physical world.

We make the following contributions:

\begin{itemize}[leftmargin=1cm, noitemsep, nosep]
\item We introduce \name, a human-intuitive interface that enables agentic VLMs to perform closed-loop robotic manipulation through discrete motion primitives without task-specific robot training. It offers a new blueprint for general embodied intelligence through \textbf{intelligence transfer} from pretrained multimodal models to embodied control.
\item We develop an interface-aligned ICL scheme that enables frozen VLMs to learn how to act from only a few demonstrations, jointly grounding primitive action semantics and task-level strategies. Remarkably, even a single demonstration can substantially improve robotic manipulation performance without any parameter updates.
\item Experiments show that \name{} achieves SOTA performance on both RoboTwin 2.0 C2R and RoboDojo with only a single in-context demonstration, surpassing prior robot-trained and agentic methods without task-specific parameter updates.
% as well as real-world cloth-folding and pick-and-place tasks on Franka and Piper platforms, demonstrating generalization across tasks, environments, and embodiments and strong performance relative to existing agentic and robot-trained approaches.
\end{itemize}

\section{Related Work}
\label{gen_inst}

\subsection{From Task-Specific Intelligence to Transferable Intelligence}
The development of artificial intelligence has shifted from task-specific competence toward increasingly transferable (or generalized) intelligence. Early deep learning systems~\citep{lecun1998gradient,krizhevsky2012imagenet} achieved remarkable performance on individual proxy tasks, such as image classification~\citep{he2016deep,simonyan2014very}, object detection~\citep{redmon2016you,ren2016faster}, and machine translation~\citep{sutskever2014sequence,bahdanau2014neural}, but the resulting capabilities were largely tied to the particular task and objective on which they were trained. The emergence of large language models (LLMs)~\citep{brown2020language,ouyang2022training} fundamentally changed this paradigm: a single pretrained model can perform and rapidly adapt to a broad range of tasks~\citep{liu2023visual,chen2021evaluating,yao2022react} through language instruction, without task-specific training.
Such generalization suggests a form of intelligence that is increasingly reusable across tasks, resembling human intelligence in its ability to transfer previously acquired perception, knowledge, learning, and reasoning capabilities to new problems. 
In this work, we focus on such transferable intelligence and ask whether intelligence acquired predominantly in the digital world can further transfer to the physical world  without specific training.

\subsection{End-to-end Robotic Policy}
End-to-end robotic policies directly predict executable actions from robot observations and task conditions. This paradigm dates back to early deep visuomotor control and large-scale robot learning~\citep{DBLP:journals/jmlr/LevineFDA16, DBLP:conf/icra/PintoG16}. With the development of transformer- and diffusion-based architectures, end-to-end policies have gradually evolved from task-specific policy learning toward increasingly generalist robot control~\citep{DBLP:conf/rss/BrohanBCCDFGHHH23, DBLP:conf/rss/GhoshWPBMDHK0LT24, DBLP:conf/iclr/LiuWLTCWX0025}, with larger and more diverse robot datasets enabling improved generalization across tasks, environments, and embodiments.
VLA models leverage pretrained VLMs and further adapt their semantic and reasoning capabilities to robot action generation~\citep{pmlr-v229-zitkovich23a, DBLP:conf/corl/KimPKXB0RFSVKBT24, DBLP:journals/corr/abs-2510-10274, DBLP:journals/corr/abs-2410-24164, DBLP:journals/corr/abs-2503-14734, DBLP:journals/corr/abs-2503-20020}. Meanwhile, WAMs, building on recent world models~\citep{DBLP:conf/iclr/YangDGTKSA24, DBLP:conf/icml/BruceDEPS0LMSAA24}, further unify future prediction and robot action generation~\citep{DBLP:journals/corr/abs-2602-15922, DBLP:journals/corr/abs-2604-27792}. 
Despite these advances, these approaches still require training on action data. In contrast, we investigate whether the intelligence already present in pretrained VLMs can be directly transferred to closed-loop robotic control through an appropriate action interface, without task-specific parameter updates.
% Despite these advances, these approaches still require robot action data to learn an executable policy. In contrast, we investigate whether the intelligence already present in pretrained VLMs can be directly transferred to closed-loop robotic control through an appropriate action interface, without task-specific parameter updates.

\subsection{Agentic Methods for Robotic Control}
An alternative line of work uses pretrained VLMs as online agents for robotic control, rather than relying solely on monolithic end-to-end policies. Earlier approaches primarily employ foundation models for high-level planning and decision making, grounding their predictions into executable robot behaviors through pretrained skill libraries, environmental feedback, or programmatic control interfaces~\citep{DBLP:conf/corl/IchterBCFHHHIIJ22, DBLP:conf/icra/LiangHXXHIFZ23,DBLP:conf/corl/HuangXXCLFZTMCS22, DBLP:conf/corl/HuangWLZF24}.
Recently, agentic robotic systems have increasingly incorporated memory, verification, replanning, and closed-loop interaction, while most still rely on pretrained VLA policies for low-level action execution~\citep{DBLP:journals/corr/abs-2505-23450,DBLP:journals/corr/abs-2604-13942,DBLP:journals/corr/abs-2603-22435,DBLP:journals/corr/abs-2607-08448}. Concurrent to our work, Show-Harness~\citep{chen2026showharnessjustvlmagent} similarly demonstrates that frontier VLMs can directly perform closed-loop robot control through a compact semantic action interface. 
Compared with Show-Harness, \name{} goes beyond designing a harness for VLMs and also focuses on developing an effective ICL strategy co-design, which significantly improves their performance.

\section{\name}
\label{method}

% ============================================================
\subsection{Overall Framework}
\label{sec:overall}

\begin{figure}[t]
\centering
\includegraphics[width=\textwidth]{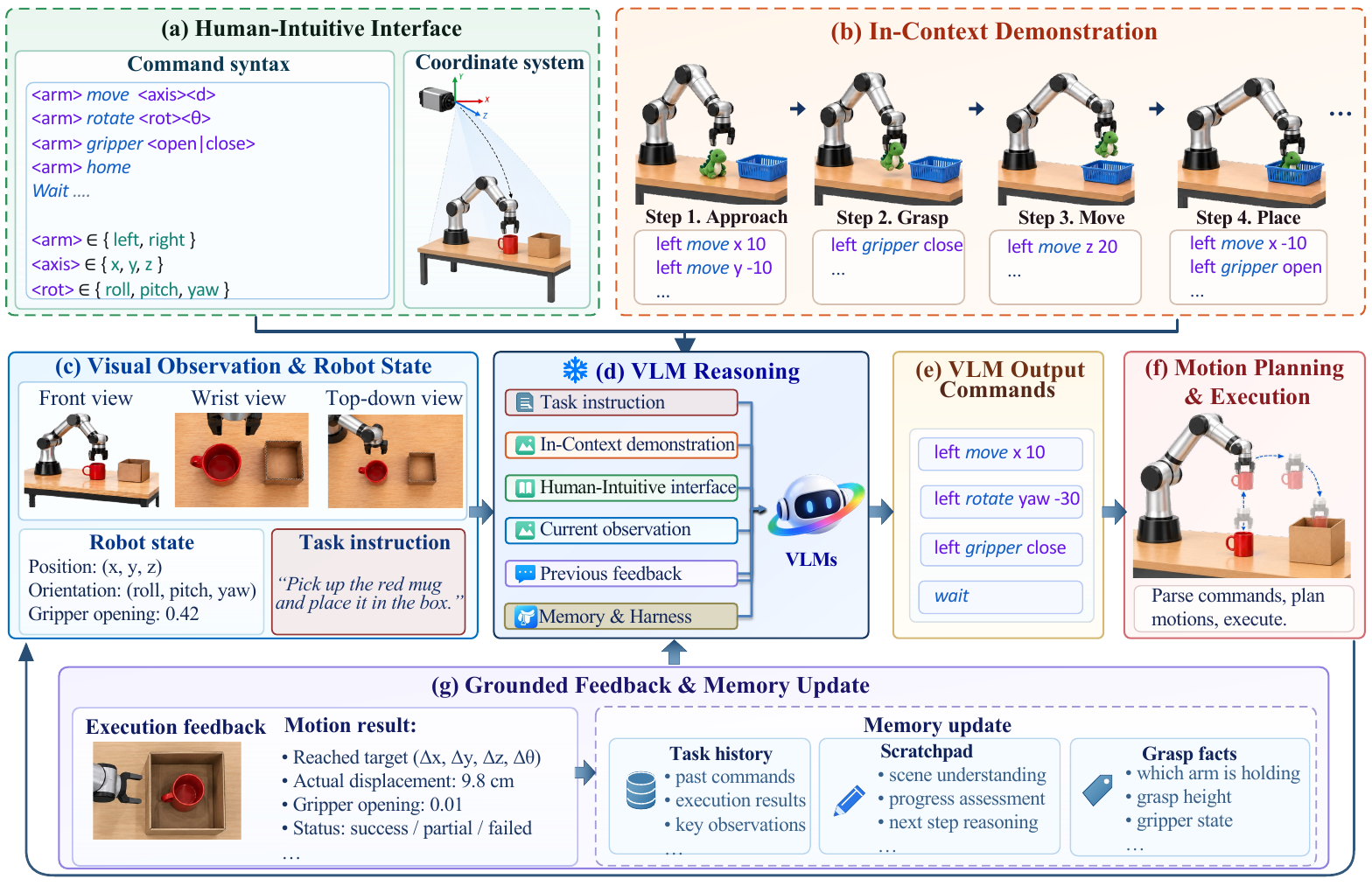}
\caption{Overview of \name{}. \name{} conditions a pretrained VLM on (a) a human-intuitive semantic action interface and (b) in-context demonstrations. At each interaction round, the VLM receives (c) multi-view visual observations and robot state together with task context, execution feedback, and memory, performs (d) reasoning, and generates (e) semantic action commands. The commands are grounded into (f) planned robot motions and executed in the environment; the resulting physical outcomes are then converted into (g) grounded feedback and memory updates for the next round. This loop enables continual observation, reasoning, execution, and adaptation without task-specific parameter updates.}
\label{fig:pipeline}
\end{figure}

Figure~\ref{fig:pipeline} provides an overview of \name.
A manipulation task can be specified by a natural-language instruction $L$.
For each task, at decision round $t$, the VLM receives annotated visual observations
$I_t$, measured robot state $x_t$, execution feedback $F_{t-1}$ from
the previous rounds, and interaction memory $M_t$.
In addition, the model is conditioned on two forms of context that
remain fixed throughout an episode:
a robot--environment profile $E$ and an in-context demonstration set
$D$.
The profile describes interface conventions such as
workspace constraints, camera, grid-based localization, and gripper properties,
whereas $D$ provides in-context examples of how the interface can be used to
interact with the environment.
The closed-loop interaction can be written as:

\begin{equation}
\label{eq:pipeline}
\begin{aligned}
    (y_t,\mathbf{a}_t)
    &= \pi_\theta
       \left(
       L,E,D;
       I_t,x_t,F_{t-1},M_t
       \right), \\
    (s_{t+1},F_t)
    &= \mathcal{E}_P
       \left(
       s_t,\mathbf{a}_t
       \right), \\
    (I_{t+1},x_{t+1})
    &= \mathcal{O}_P(s_{t+1}), \\
    M_{t+1}
    &= \mathcal{U}
       \left(
       M_t,
       \mathbf{a}_t,
       y_t,
       F_t,
       x_{t+1}
       \right).
\end{aligned}
\end{equation}

Here, $\pi_\theta$ denotes a pretrained VLM whose parameters remain
frozen.
The model outputs a 
sequence of semantic action commands
$\mathbf{a}_t$ together with a structured response $y_t$. 
The structured response $y_t$ contains the estimate of task progress, the current plan, and a compact
scratchpad.
The execution operator $\mathcal{E}_P$ parses these commands, grounds
them into robot motions, executes them, and converts the resulting
physical outcome into feedback $F_t$.
The observation operator $\mathcal{O}_P$ then constructs the next
visual and proprioceptive observation, while $\mathcal{U}$ updates
the interaction memory from both model-generated and
execution-derived information.
The environment state $s_t$ is introduced only to formalize physical
state transitions and is not directly exposed to the VLM.
In particular, online control does not rely on privileged object poses.
Throughout an episode, $\theta$, $E$, and $D$ are fixed;
adaptation arises from the evolving observation, execution feedback,
and memory.

% ============================================================
\subsection{Human-Intuitive Interface}
\label{sec:interface}

The first component of \name{} is a human-intuitive interface that
bridges multimodal reasoning and physical robot control.
Our goal is to expose robotic interaction in a form that is both easy
for a pretrained VLM to interpret and straightforward to ground into
physical execution.
We firstly define the \emph{gripper interaction point} (GIP) as the midpoint between
the two fingertips of the gripper, which directly represents the point
at which the robot interacts with objects.
We use this GIP consistently for visual annotations, robot-state
reporting, and motion commands, avoiding ambiguity between the wrist
pose used by the low-level controller and the actual interaction point
of the gripper.

% \paragraph{VLM-readable observations.}
% Raw RGB observations alone can make metric spatial reasoning difficult,
% especially when the model must estimate planar displacement, height,
% and gripper--object alignment.
% We therefore combine multiple complementary views with lightweight
% geometric annotations.
% Specifically, the visual observation $I_t$ contains a global
% perspective view, a top-down view, and wrist-mounted views.
% Using camera calibration, we project the GIP locations, a metric
% tabletop grid, and coordinate-axis legends onto the images.
% The top-down view primarily supports planar localization,
% the global view provides scene and height context,
% and wrist views support fine alignment near grasping points.

% The visual observation is accompanied by measured robot state $x_t$.
% For each arm, we provide the GIP position, its approach and finger
% axes, and the measured gripper opening, together with the table height
% and interaction progress.
% These measurements describe the robot and its reference geometry,
% rather than privileged object states:
% object locations and relations must still be inferred from visual
% observations.

\paragraph{Semantic action interface.}
We adopt a game-like action interface built from simple, semantically meaningful spatial primitives. Such commands are intuitive to humans and resemble interaction abstractions commonly used in games, teleoperation, and instructional content, making them more likely to align with concepts already encountered during web-scale VLM pretraining. As shown in Fig.~\ref{fig:pipeline}(a), instead of directly predicting joint-level actions or high-frequency continuous end-effector controls, the VLM interacts with the robot through a compact vocabulary of parameterized semantic commands.
The complete command grammar is
\begin{equation}
\label{eq:action_space}
\begin{aligned}
\mathcal{A}=\{&
\texttt{<arm> move <axis> <d>},\\
&
\texttt{<arm> rotate <rot> <$\theta$>},\\
&
\texttt{<arm> point <pose>},\\
&
\texttt{<arm> gripper <g>},\\
&
\texttt{<arm> home},~
\texttt{wait},~
\texttt{done}
\}.
\end{aligned}
\end{equation}
Here,
$\texttt{arm}\in\{\texttt{left},\texttt{right}\}$;
translation axes are
$\{\texttt{x},\texttt{y},\texttt{z}\}$ and
rotation axes are
$\{\texttt{roll},\texttt{pitch},\texttt{yaw}\}$,
both referring to the world frame;
orientation presets include
\texttt{down}, \texttt{forward}, and \texttt{down45};
and the gripper argument is \texttt{open}, \texttt{close}, or a
normalized opening in $[0,1]$.
All spatial commands are defined with respect to the GIP and specify
incremental changes of its pose:
\texttt{move} translates the GIP by $d$ along the given axis while
preserving its orientation, whereas \texttt{rotate} turns it by
$\theta$ about the given axis while preserving its position.
Translation and rotation magnitudes are clipped to $20$\,cm and
$90^\circ$ per command, respectively.
The \texttt{point} command provides several common gripper-orientation
presets, while \texttt{home}, \texttt{wait}, and \texttt{done}
provide simple high-level utilities for resetting an arm, allowing
the environment to settle, and requesting task-completion checking.

\paragraph{From semantic commands to physical motion.}
Each semantic motion command is translated into a complete planned motion to a target GIP pose and executed until the robot reaches a stationary state, abstracting away low-level trajectory generation and control from the VLM. 

% The VLM operates at a lower decision frequency and influences high-level action commands, reducing inference calls in free space while allowing more frequent re-observation and correction near objects.
% A semantic motion command is not interpreted as one low-level control
% increment.
% Instead, the interface first converts it into a target GIP pose,
% maps this target to the wrist-pose convention required by the robot,
% and invokes the low-level motion planner.
% The resulting trajectory is executed until the robot reaches a
% stationary state.
% Thus, one semantic command corresponds to one complete planned motion.

% This temporal abstraction separates the roles of the VLM and the
% robot controller:
% the VLM reasons over semantically meaningful state changes at a
% relatively low decision frequency, while trajectory generation,
% collision-aware motion, and high-frequency control are delegated to
% the underlying robot stack.
% At each decision round, the VLM may output a short action chunk of up
% to four commands, which are executed sequentially before the next
% observation is acquired.
% For free-space motion, this reduces expensive VLM calls;
% near objects, the profile encourages shorter action chunks so that the
% agent can re-observe the scene and correct its behavior more frequently.

% ============================================================
\subsection{In-Context Learning Design}
\label{sec:icl}

A simple interface together with language-based action rules, however, does not eliminate all ambiguity~\cite{yu2022using}.
The model must still understand the semantics and granularity of individual actions, infer how each action changes the observed scene, and learn interaction conventions that may not be obvious from the command names alone. Much like humans, who may struggle to fully understand unfamiliar rules from written instructions but can grasp them more easily after observing a few examples, VLMs can benefit substantially from in-context demonstrations. This is particularly useful for action concepts such as gripper rotations and orientation changes, which may be less explicitly represented in web-scale pretraining data than simple translational motions; demonstrations provide direct visual examples that help the VLM learn their effects and incorporate them into task-level reasoning.
We decompose the demonstration context as
\begin{equation}
\label{eq:demo_context}
    D
    =
    D_{\mathrm{prim}}
    \oplus
    D_{\mathrm{task}},
\end{equation}
where $D_{\mathrm{prim}}$ is a shared command primer illustrating all
basic command effects, and
\begin{equation}
    D_{\mathrm{task}}
    =
    \{
        \mathcal{D}^{(m)}
    \}_{m=1}^{N_D}
\end{equation}
contains task-level demonstrations.
This formulation does not assume a fixed number of examples:
$N_D=0$, $N_D=1$, and $N_D>1$ correspond to zero-shot,
one-shot, and few-shot settings. 
Together, \(D_{\mathrm{prim}}\) and \(D_{\mathrm{task}}\) provide a two-level demonstration context: the former establishes how each primitive command affects the robot, while the latter shows when and how these primitives are composed into complete task-solving behaviors.

\paragraph{Interface-compatible demonstrations.}
Raw expert trajectories are not directly suitable as in-context
examples, since they consist of continuous low-level actions that
differ from those available to \name.
We therefore express each trajectory in the semantic command space of
Sec.~\ref{sec:interface}.
A trajectory is first reduced to a sequence of end-effector waypoints
and gripper states.
Each waypoint is then reached with a short sequence of translation,
rotation, and gripper commands, which turns the entire trajectory into
a command sequence that the online model could have issued itself.

\paragraph{Complete in-context demonstrations.}
Demonstrations are collected in scenes disjoint from those used for
evaluation, from the benchmark's scripted expert in simulation and
from human teleoperation on real robots;
in simulation, these are the same trajectories used to train the
robot policies we compare against.
Each demonstration consists of $N_m$ interaction rounds,
\begin{equation}
\label{eq:demonstration}
    \mathcal{D}^{(m)}
    =
    \left\{
    \left(
        I_j^{(m)},
        x_j^{(m)},
        r_j^{(m)},
        \mathbf{a}_j^{(m)},
        f_j^{(m)}
    \right)
    \right\}_{j=1}^{N_m},
\end{equation}
where $I_j^{(m)}$ is the visual observation at round $j$,
$x_j^{(m)}$ the robot state,
$\mathbf{a}_j^{(m)}$ the commands issued,
$f_j^{(m)}$ their physical effect, and
$r_j^{(m)}$ a short rationale.
The physical effect $f_j^{(m)}$ is derived from consecutive states as
the change in GIP pose and gripper opening.
Note that $I_j^{(m)}$ may be empty for some rounds, 
while retaining the full textual trajectory for planning guidance.
This sparsification is applied to long-horizon RoboDojo trajectories, where the number of rounds can substantially exceed the in-context image budget. We set the upper bound on the in-context image budget to 16 observations per round. 
When this budget is exceeded, we retain images only for semantically informative rounds, such as grasping, rotation, and task completion, while omitting visually redundant transition rounds.
The rationale $r_j^{(m)}$ is written afterwards by a VLM, which
reviews the recorded episode with a task-agnostic prompt and states,
for each round, the plan behind its commands in the same format as
the responses of online model.
% Since a demonstration is recorded in a different scene from the one
% being evaluated, the online model is instructed to follow the
% demonstrated strategy while acting according to its own scene.

% We select only a small number of informative keyframes,
% prioritizing the start and end of the trajectory and rounds involving
% important interaction events such as gripper commands or orientation
% changes.
% Commands that produce essentially no motion are removed,
% whereas partially executed commands are retained together with their
% actual effect to preserve consistency between images and actions.

% The scripted experts provide actions but only templated textual plans.
% We therefore use a VLM offline to rewrite the scene, progress, and plan
% descriptions while keeping the recorded images and commands unchanged.
% This produces rationales that are more natural for a VLM to interpret,
% without treating them as ground-truth expert reasoning.
% The resulting demonstration is fixed throughout the episode.
% The online model is explicitly instructed to transfer the demonstrated
% strategy rather than copy numerical positions, arm choices, or exact
% motion magnitudes from the example. 

% ============================================================

\section{Experiments}

 We evaluate \name{} in both simulated and real-world manipulation settings. In simulation, we consider two challenging benchmarks RoboTwin 2.0~\citep{chen2025robotwin}, and RoboDojo~\citep{chen2026robodojo} covering diverse tasks and manipulation capabilities. We further deploy \name{} on real robots to evaluate its transfer across embodiments and physical environments.

\begin{table}[t]
\centering
\setlength{\tabcolsep}{15pt} % 调整列间距，默认约为 6pt
\caption{
Evaluation on RoboTwin 2.0 C2R~\cite{chen2025robotwin}.
Full Set follows the official training protocol, where a single policy is jointly post-trained on all 50 benchmark tasks using 50 clean demonstrations per task. 1-Shot denotes using one randomly sampled demonstration from the clean set. Success rates (\%) are reported under the domain-randomized evaluation setting. CC denotes Claude Code. HarnessVLA uses the pretrained LingBot-VLA as the default VLA.
}
\label{tab:robotwin2}
\begin{tabular}{lcc}
\toprule
\textbf{Method} & \textbf{Number of Shots} & \textbf{Success rate (\%)}\\
\midrule

FastWAM~\cite{yuan2026fast}
& Full Set
& 1.9 \\

StarVLA ~\cite{community2026starvla}
& Full Set
& 10.6 \\

GalaxeaVLA~\cite{jiang2025galaxea}
& Full Set
& 12.7 \\

Xiaomi-Robotics-0~\cite{cai2026xiaomi}
& Full Set
& 18.2
 \\

GR00T-1.7\cite{DBLP:journals/corr/abs-2503-14734}
& Full Set
& 20.7
 \\

X-VLA~\cite{DBLP:journals/corr/abs-2510-10274}
& Full Set
& 20.9
 \\

X-WAM~\cite{guo2026unified}
& Full Set
& 25.8
 \\

$\pi_{0.5}$~\cite{DBLP:journals/corr/abs-2504-16054}
& Full Set
& 46.0
 \\

LingBot-VLA~\cite{wu2026pragmatic}
& Full Set
& 50.4

 \\
\midrule
HarnessVLA (Codex)~\cite{DBLP:journals/corr/abs-2607-08448}
& Full Set
& 58.0
 \\

HarnessVLA (CC)~\cite{DBLP:journals/corr/abs-2607-08448}
& Full Set
& 58.4
 \\

\midrule
\rowcolor{green!7}
\name{}(Ours, Gemini-3.8-Flash) 
& 0
& 47.0
 \\

\rowcolor{green!7}
\name{}(Ours, Gemini-3.8-Flash) 
& 1
& 62.2
 \\

% \rowcolor{green!7}
% \name{}(Ours, Gemini-3.8-Flash) 
% & 2
% & 63.6
%  \\

% \rowcolor{green!7}
% \name{}(Ours, Gemini-3.8-Flash) 
% & 4
% & 65.4
%  \\

\rowcolor{green!7}
\name{}(Ours, GPT-6 Astra) 
& 0
& 53.2
 \\

\rowcolor{green!7}
\name{}(Ours, GPT-6 Astra) 
& 1
& \textbf{73.6}
 \\

% \rowcolor{green!7}
% \name{}(Ours, Gemini-Flash 3.8) 
% & 8
% & XX.X
%  \\

\bottomrule
\end{tabular}
\end{table}

\subsection{Simulation Benchmark Test}

\subsubsection{RoboTwin 2.0}
We adopt the C2R setting of RoboTwin 2.0, which contains 50 bimanual manipulation tasks. Following the official C2R protocol, robot-trained baselines are jointly post-trained on 50 clean demonstrations per task and evaluated in domain-randomized environments. 
\name{} does not train on these demonstrations; instead, we evaluate zero-shot and few-shot variants, where task demonstrations are provided only as in-context examples in clean set to avoid data leakage.
We report the average task success rate across the benchmark. This C2R setting specifically evaluates generalization from clean scenes (one-shot from here) to randomized scenes. For each task, we conduct 10 independent evaluation runs.
% \TODO{explain 10 run here.}

Tab.~\ref{tab:robotwin2} compares \name{} with representative robot-trained and agentic baselines on RoboTwin 2.0 C2R.
While prior methods are trained or post-trained on the full benchmark training set, \name{} performs robot control without task-specific parameter updates and uses at most a single demonstration as in-context guidance. Even in the zero-shot setting, \name{} with GPT-6 Astra achieves a 53.2\% success rate, outperforming several strong robot policies including $\pi_{0.5}$ (46.0\%) and LingBot-VLA (50.4\%). With only one in-context demonstration, its success rate further increases to 73.6\%, exceeding the SOTA agentic VLA method, HarnessVLA (Claude Code), by 15.2 percentage points.
These results highlight that a pretrained VLM can effectively translate its general-purpose reasoning capabilities into robotic manipulation through a lightweight action interface, while benefiting substantially from few-shot in-context adaptation.

\paragraph{Ablation Study.}
We conduct ablation studies on RoboTwin 2.0 to examine the effects of the number of in-context demonstrations, the choice of VLM, and key harness components. 
As shown in Tab.~\ref{tab:ablation}, first, increasing the number of demonstrations generally improves performance: with Gemini-3.8-Flash, the success rate rises from 47.0\% in the zero-shot setting to 62.2\% with one demonstration, and further reaches 65.4\% with four demonstrations. 
The marginal gain becomes smaller beyond one shot, while performance slightly drops to 62.7\% with eight demonstrations, suggesting that excessively long context may interfere with effective VLM understanding, which is a commonly observed challenge in long-context VLM reasoning~\cite{ge2025v2pe}.
Second, \name{} benefits substantially from stronger VLMs. Under the same one-shot setting, the success rate increases from 14.4\% with GPT-5.6-Luna to 43.2\% with GPT-5.6-Sol, 62.2\% with Gemini-3.8-Flash, and 73.6\% with GPT-6 Astra, indicating that the effectiveness of \name{} scales with the underlying model capability. 
Finally, we ablate several components of the harness under the zero-shot setting. Removing reasoning reduces the success rate from 47.0\% to 34.8\%, while removing grid-based localization causes a larger drop to 32.4\%. Removing the command primer leads to a smaller but still noticeable decrease to 44.0\%. These results show that explicit reasoning and spatial grounding are particularly important for reliable robot control, while the command primer provides additional guidance for interpreting the action interface.

% \begin{wrapfigure}{r}{0.5\textwidth}
%   \centering
%   \includegraphics[width=0.5\textwidth]{images/robodojo.pdf}
%   \caption{robodojo}
%   \label{fig:robodojo}
% \end{wrapfigure}
\begin{figure*}[t]
    \centering

    % ================= Left: Table =================
    \begin{minipage}[t]{0.5\textwidth}
        \vspace{0pt}
        \captionof{table}{
            Inference and execution efficiency of representative
            VLA and WAM policies.
            \textbf{Infer.}: inference latency;
            \textbf{Motion}: motion execution time;
            \textbf{I/M}: inference-to-motion time ratio.
Seed-2.1-Pro on Volcano Engine is used as the VLM of \name{}, considering network latency. cmds denotes commands. }
        \label{tab:efficiency}

        \vspace{0.5em}

        \centering
        \small
        \setlength{\tabcolsep}{2pt}
        \renewcommand{\arraystretch}{1.05}

        \resizebox{\linewidth}{!}{%
        \begin{tabular}{lcccc}
            \toprule
            \textbf{Model} &
            \textbf{Infer.} &
            \textbf{Actions} &
            \textbf{Motion} &
            \textbf{I/M} \\
            \midrule
            $\pi_{0.5}$ & 101 ms & 45.0 steps & 2.70 s & 0.037 \\
            StarVLA     & 70 ms  & 16.0 steps & 0.96 s & 0.073 \\
            X-VLA       & 143 ms & 28.6 steps & 1.71 s & 0.084 \\
            \midrule
            FastWAM     & 523 ms & 27.3 steps & 1.64 s & 0.32 \\
            Motus       & 1.93 s & 16.0 steps & 0.96 s & 2.01 \\
            LingBot-VA  & 8.89 s & 22.2 steps & 1.33 s & 6.67 \\
            \midrule
            \name{}     & 9.74 s & 3.4 cmds & 2.09 s & 4.65 \\
            \bottomrule
        \end{tabular}%
        }
    \end{minipage}
    \hfill
    % ================= Right: Figure =================
    \begin{minipage}[t]{0.46\textwidth}
        \vspace{0pt}
        \centering

        \includegraphics[
            width=\linewidth
        ]{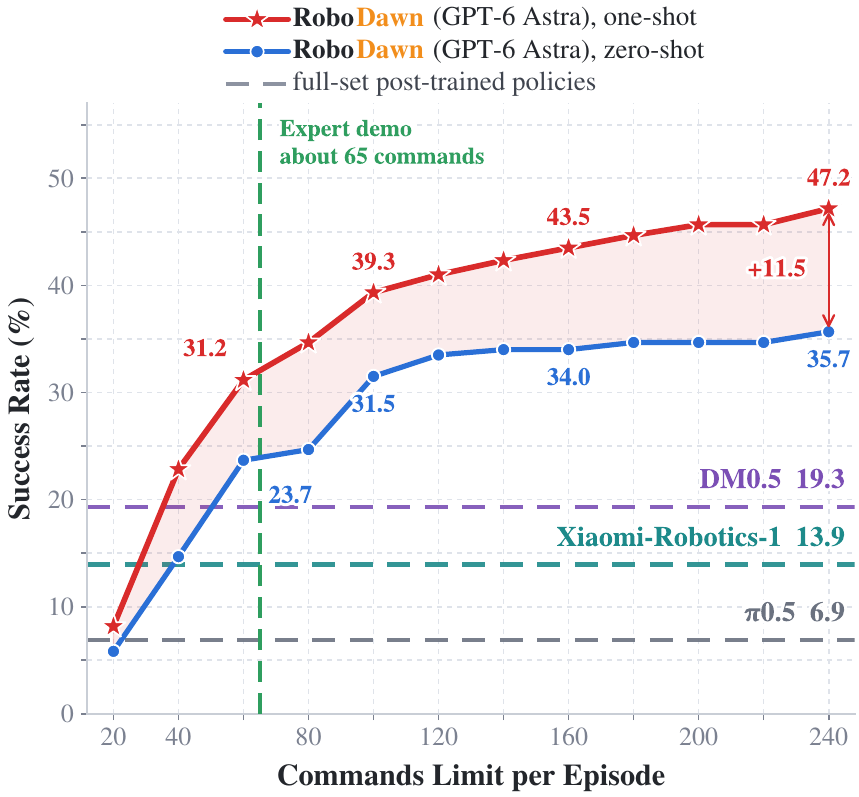}

        \captionof{figure}{RoboDojo success rate under varying per-episode command budgets. \name{} exhibits clear test-time scaling.
        % benefiting from its strong error-recovery and memory capabilities. 
        }
        % RoboDawn (GPT-6 Astra) is evaluated in one-shot and zero-shot settings. The green dashed line denotes the expert demonstration budget ($\sim$65 commands), and horizontal dashed lines denote full-set post-trained policies. At 240 commands, one-shot reaches 46.5\%, outperforming zero-shot by 13.2 percentage points.}
        \label{fig:robodojo}
    \end{minipage}

\end{figure*}

\paragraph{Execution Efficiency.}
To evaluate efficiency, we run each of the 50 tasks in RoboTwin 2.0 and report the average computational cost.
Considering practical factors such as network latency, we use Seed-2.1-Pro as the VLM for latency evaluation. Tab.~\ref{tab:efficiency} compares the execution efficiency of \name{} with representative VLA and WAM policies.
Beyond reporting inference latency alone, we additionally measure the inference-to-motion ratio to characterize the potential for pipelined execution between model reasoning and physical robot motion. In particular, when this ratio is below 0.5, inference can potentially be sufficiently overlapped with motion execution to enable streaming inference.

% \begin{table}[t]
%     \centering
%     \caption{Inference and execution efficiency comparison with representative VLA and WAM policies. Considering factors such as network latency, we chose Seed-2.1-Pro on Volcano Engine as the VLM.}
%     \label{tab:efficiency}
%     \begin{tabular}{lcccc}
%         \toprule
%         \textbf{Model} &
%         \textbf{Inference Time} &
%         \textbf{Actions} &
%         \textbf{Motion Time} &
%         \textbf{Inference / Motion} \\
%         \midrule
%         $\pi_{0.5}$   & 101 ms & 45.0 steps & 2.70 s & 0.037 \\
%         StarVLA       & 70 ms  & 16.0 steps & 0.96 s & 0.073 \\
%         X-VLA         & 143 ms & 28.6 steps & 1.71 s & 0.084 \\
%         \midrule
%         FastWAM       & 523 ms & 27.3 steps & 1.64 s & 0.32 \\
%         Motus         & 1.93 s & 16.0 steps & 0.96 s & 2.01 \\
%         LingBot-VA    & 8.89 s & 22.2 steps & 1.33 s & 6.67 \\
%         \midrule
%         \name{}       & 9.74 s & 3.4 cmds   & 2.09 s & 4.65 \\
%         \bottomrule
%     \end{tabular}
% \end{table}

\subsubsection{RoboDojo}

We additionally evaluate on RoboDojo, a comprehensive benchmark designed to test generalist manipulation across generalization, memory, long-horizon, precision, and open capabilities. Its simulation suite contains 42 tasks on a bimanual robot platform. Following the benchmark protocol, we report both success rate, which measures complete task execution, and task progress score, which additionally captures partial completion. 
Here, we report the average results over 5 runs.

As shown in Tab.~\ref{tab:robodojo}, \name{} achieves strong performance on RoboDojo without task-specific post-training. In the zero-shot setting, it attains a success rate of 35.67\%. Providing only a single demonstration further improves the success rate to 47.17\%. This result substantially surpasses full-set post-trained policies such as DM0.5, which achieves a success rate of 19.34\%.

Beyond final task performance, Fig.~\ref{fig:robodojo} reveals a clear test-time scaling behavior: as the command budget increases, the success rate consistently improves in both the zero-shot and one-shot settings. In particular, the one-shot success rate increases from 31.2\% with 60 commands to 47.2\% with 240 commands, while the zero-shot variant improves from 23.7\% to 35.7\%. This consistent scaling suggests that RoboDawn can effectively translate additional test-time computation and interaction into improved task success. We attribute this behavior to \name{}'s strong failure-recovery capability and memory mechanism, which enable it to adapt its subsequent actions based on previous observations and execution outcomes.

\subsection{Real-World Robotics Deployment}

\textbf{Real-world evaluation.} We further deploy \name{} with Gemini 3.8 Flash on real robots to evaluate whether the proposed harness transfers beyond simulation under zero-shot setting.
As shown in Tab.~\ref{tab:real-world}, \name{} achieves promising performance on real-world block-in-basket and block-stacking tasks, demonstrating that the same harness can be directly applied to physical robots in zero-shot setting.
Block stacking achieves a lower success rate than block-in-basket, mainly because it requires more precise perception and spatial alignment during placement.
Cloth folding remains considerably more challenging.
We hypothesize that this is mainly because the task involves substantial end-effector rotation and orientation adjustment, which may be less frequently represented in the web-scale data used to train VLMs and are therefore harder for the model to reason about reliably.
In fact, we observed several near-successful cases in the cloth-folding task.
However, the final folds were not sufficiently neat, so we did not count these trials as successful.

\begin{table*}[t]
    \centering
    \renewcommand{\arraystretch}{1.05}
    \captionsetup[subtable]{
        justification=centering,
        singlelinecheck=true
    }
    \caption{Ablation studies of \name{} on RoboTwin 2.0.}

    % ==================== (a) Shots ====================
    \begin{subtable}[t]{0.26\textwidth}
        \centering
        \caption{\textbf{Number of shots (Gemini-3.8-Flash).}}
        \begin{tabular*}{\linewidth}{@{\extracolsep{\fill}}cc@{}}
            \toprule
            \# Shots & SR (\%) \\
            \midrule
            0 & 47.0 \\
            1 & 62.2 \\
            2 & 63.6 \\
            \rowcolor{gray!20}
            4 & \textbf{65.4} \\
            8 & 62.7 \\
            \bottomrule
        \end{tabular*}
        
        \label{tab:abl-shots}
    \end{subtable}
    \hfill
    % ==================== (b) Models ====================
    \begin{subtable}[t]{0.32\textwidth}
        \centering
        \caption{\textbf{Models. Results use 1-shot by default}}
        \begin{tabular*}{\linewidth}{@{\extracolsep{\fill}}cc@{}}
            \toprule
            Model & SR (\%) \\
            \midrule
            GPT-5.6-Luna & 14.4 \\
            % Qwen3.8-Max & 36.6 \\
            GPT-5.6-Sol & 43.2 \\
            Seed-2.1-Pro & 45.0 \\
            Gemini-3.8-Flash & 62.2 \\
            \rowcolor{gray!20}
            GPT-6 Astra & \textbf{73.6} \\
            \bottomrule
        \end{tabular*}
        
        \label{tab:abl-models}
    \end{subtable}
    \hfill
    % ==================== (c) Harness ====================
    \begin{subtable}[t]{0.32\textwidth}
        \centering
        \caption{\textbf{Harness (Gemini 3.8 Flash, zero-shot).)}}
        \begin{tabular*}{\linewidth}{@{\extracolsep{\fill}}cc@{}}
            \toprule
            Harness & SR (\%) \\
            \midrule
            \rowcolor{gray!20}
            Full & \textbf{47.0} \\
            w/o reasoning & 34.8 \\
            w/o grids & 32.4 \\
            w/o primer & 44.0 \\
            \bottomrule
        \end{tabular*}
        
        \label{tab:abl-harness}
    \end{subtable}
    \label{tab:ablation}
\end{table*}

\begin{table}[t]
\centering
\setlength{\tabcolsep}{10pt} % 调整列间距，默认约为 6pt
\caption{
Evaluation on RoboDojo~\cite{chen2026robodojo}.
Full Set follows the official RoboDojo training protocol, in which a single policy is post-trained on the full RoboDojo-specific training set. Score denotes the task progress score. 
Compared with the full set of 42 tasks, the 34 tasks setting excludes 8 open tasks. For these 8 tasks, other methods may not have corresponding training data. } 
\label{tab:robodojo}
\begin{tabular}{lccc}
\toprule
\textbf{Method} & \textbf{\# Shots} & \textbf{Score} & \textbf{Success rate (\%)}
\\
\midrule

LingBot-VLA~\cite{wu2026pragmatic}
& Full Set
& 5.50
& 2.96 \\

X-VLA~\cite{DBLP:journals/corr/abs-2510-10274}
& Full Set
& 10.13
& 6.52
 \\

$\pi_{0.5}$~\cite{DBLP:journals/corr/abs-2504-16054}
& Full Set
& 11.41
& 6.91
 \\

Hy-Embodied-0.5-VLA~\cite{zhang2026hy}
& Full Set
& 13.07
& 8.80 \\

Xiaomi-Robotics-1~\cite{xiaomir1}
& Full Set
& 20.07
& 13.93\\

GalaxeaVLA (G0.5)~\cite{liu2026g0}
& Full Set
& 20.23
& 14.88\\

DM0.5~\cite{dm05}
& Full Set
& 24.90
& 19.34 \\

GPT-6 Astra~\cite{zhang2026gpt6astrarobodojo}
& 0
& 28.97
& 22.58 \\

\midrule
\rowcolor{green!7}
\name{}(Ours, GPT-6 Astra, all tasks) 
& 0
& 39.92
& 35.67
\\

\rowcolor{green!7}
\name{}(Ours, GPT-6 Astra, all tasks) 
& 1
& \textbf{54.63}
& \textbf{47.17}
\\

\midrule
\midrule
\rowcolor{yellow!7}
\name{}(Ours, GPT-6 Astra, 34 tasks) 
& 0
& 39.02
& 33.96
\\

\rowcolor{yellow!7}
\name{}(Ours, GPT-6 Astra, 34 tasks) 
& 1
& \textbf{51.29}
& \textbf{43.33}
\\

% \midrule

% \rowcolor{green!7}
% \name{}(Ours, GPT-6 Astra, 19 tasks) 
% & 0
% & 30.53
% & 26.32 \\

% \rowcolor{green!7}
% \name{}(Ours, GPT-6 Astra, 19 tasks) 
% & 1
% & XX.X
% & XX.X
%  \\

% \rowcolor{green!7}
% \name{}(Ours, Gemini-Flash 3.8) 
% & 8
% & XX.X
%  \\

\bottomrule
\end{tabular}
\end{table}

\begin{figure}[t]
\centering
\includegraphics[width=\textwidth]{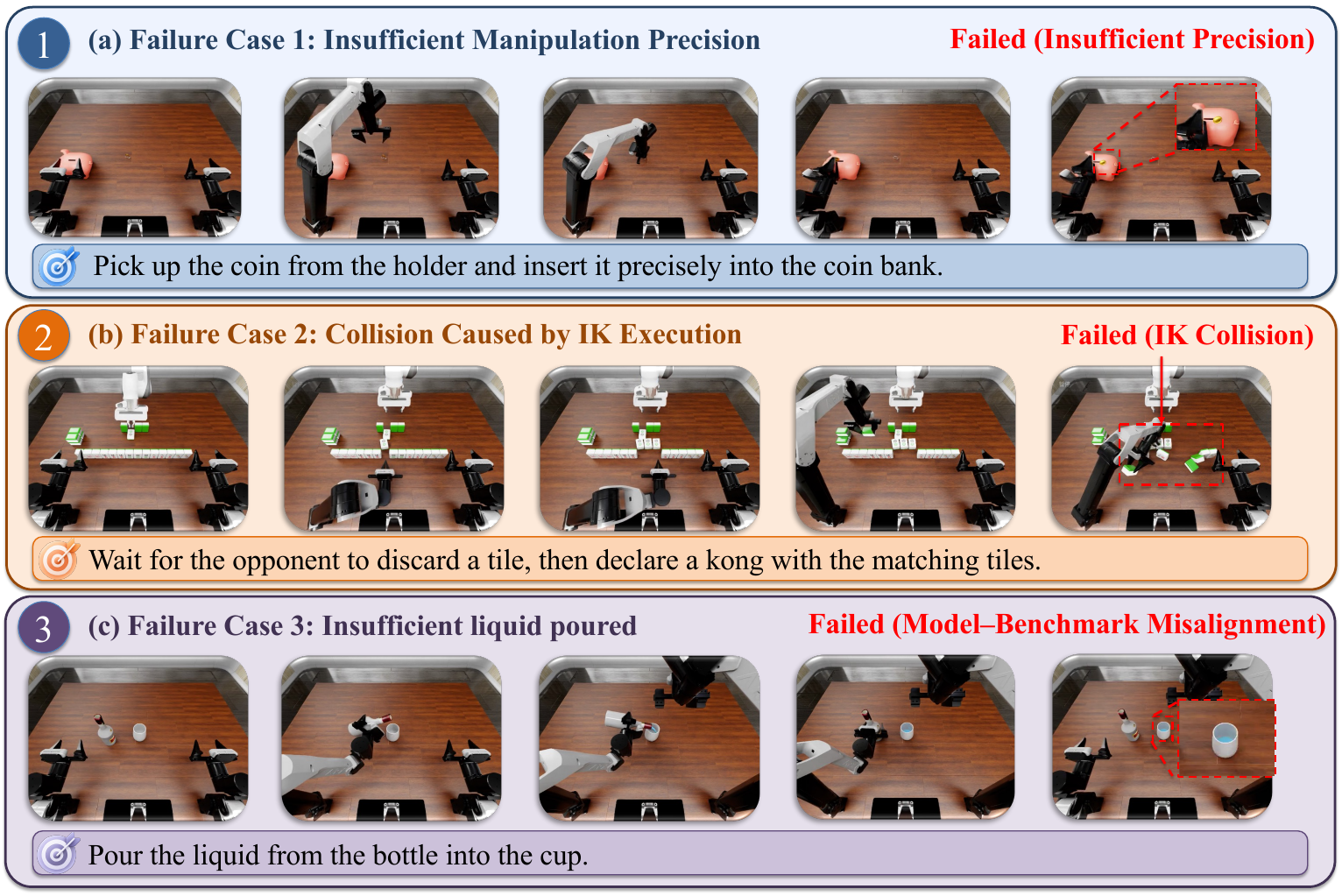}
\caption{Representative failure cases of \name{} on RoboDojo.}
\label{fig:failed_case}
\end{figure}

\begin{table}[t]
    \centering
    \setlength{\tabcolsep}{2pt}
        \caption{\textbf{Real-world evaluation.}
    Success rates on three real-world manipulation tasks.}
    \renewcommand{\arraystretch}{1.05}
    \begin{tabular}{lccc}
        \toprule
        & Block in Basket (Franka) & Block Stacking (Franka)  & Cloth Folding (Piper)  \\
        \midrule
        Success Rate 
        & 9/10
        & 5/10 
        & 0/10 \\
        \bottomrule
    \end{tabular}
    \label{tab:real-world}
\end{table}

\subsection{Failure Analysis.}
We identify three representative failure modes in RoboDojo, as illustrated in Fig.~\ref{fig:failed_case}. (a) Insufficient manipulation precision. Although the model can infer the correct high-level intention, it may fail at the final stage when the task requires very accurate positioning, such as inserting an object into a small target region. This reflects the limited fine-grained control of the current semantic action interface.
(b) mode comes from IK-related execution errors. In these cases, the predicted action sequence is reasonable at the semantic level, but the realized robot motion collides with nearby objects during execution, leading to scene disturbance and eventual failure. This reveals a mismatch between high-level semantic planning and low-level physically safe execution.
(c) mode is incorrect success judgment. The model pours liquid into the cup but stops before a sufficient amount is transferred. This reveals a misalignment between the model’s notion of task completion and the benchmark’s success criterion.

\section{Limitations}
Rome was not built in a day. This work deliberately focuses on generalization: rather than providing a perfect solution to robotic manipulation, we view \name{} as an early glimpse of a potentially more generalizable path toward embodied intelligence. Nevertheless, several important limitations remain.
\textbf{(a) Inference efficiency.} \name{} is slower than end-to-end VLA and WAM policies due to its iterative VLM reasoning and closed-loop interaction, making it less suitable for scenarios that demand high-frequency control.
\textbf{(b)Uneven difficulty across degrees of freedom (DoF).} VLMs handle translational motions more reliably than precise arm and end-effector rotations, possibly because rotational motions and 3D orientation changes are less explicitly represented in web-scale pretraining. ICL alleviates this issue to some extent, but does not fully close the gap.
\textbf{(c)Fine-grained interaction control.} \name{} remains less reliable when object interaction requires high spatial precision, particularly during fine adjustments near contact, grasping, and placement. Its discrete semantic action interface is effective for high-level manipulation, but can be too coarse for interactions that require subtle and precise corrections.
\textbf{(d)Safety in deployment.} As robotics continues to advance toward increasingly general-purpose systems, safety becomes ever more critical. Erroneous actions may cause collisions, object damage, or risks to humans.

% \textbf{Dependence on VLM capability.} \name{}’s performance remains sensitive to the underlying VLM, with substantial gaps across different models. In future, we will investigate how to make intelligence transfer more robust to model scale and backbone choice.

\section{Conclusion}

We introduce \name, a framework that transfers pretrained VLM intelligence to robotic control through a human-intuitive action interface and in-context demonstrations, without task-specific robot training. Experiments on RoboTwin 2.0, RoboDojo, and real robots show strong zero-shot performance and substantial gains from only a single demonstration. Our results suggest that general-purpose embodied intelligence may emerge not only from scaling robot data, but also from better interfaces and lessons that unlock and transfer capabilities already present in foundation models.

\bibliography{iclr2027_conference}
\bibliographystyle{iclr2027_conference}

% \appendix
% \section{Appendix}

\end{document}

%% file: math_commands.tex
\usepackage{amsmath,amsfonts,bm}

\def\eqref#1{equation~\ref{#1}}
\def\1{\bm{1}}

\DeclareMathAlphabet{\mathsfit}{\encodingdefault}{\sfdefault}{m}{sl}
\SetMathAlphabet{\mathsfit}{bold}{\encodingdefault}{\sfdefault}{bx}{n}

%% file: iclr2027_conference.bib
@article{lecun1998gradient,
  title={Gradient-based learning applied to document recognition},
  author={LeCun, Yann and Bottou, L{\'e}on and Bengio, Yoshua and Haffner, Patrick},
  journal={Proceedings of the IEEE},
  volume={86},
  number={11},
  pages={2278--2324},
  year={1998},
  publisher={Ieee}
}

@article{krizhevsky2012imagenet,
  title={Imagenet classification with deep convolutional neural networks},
  author={Krizhevsky, Alex and Sutskever, Ilya and Hinton, Geoffrey E},
  journal={Advances in neural information processing systems},
  volume={25},
  year={2012}
}

@inproceedings{he2016deep,
  title={Deep residual learning for image recognition},
  author={He, Kaiming and Zhang, Xiangyu and Ren, Shaoqing and Sun, Jian},
  booktitle={Proceedings of the IEEE conference on computer vision and pattern recognition},
  pages={770--778},
  year={2016}
}

@article{simonyan2014very,
  title={Very deep convolutional networks for large-scale image recognition},
  author={Simonyan, Karen and Zisserman, Andrew},
  journal={arXiv preprint arXiv:1409.1556},
  year={2014}
}

@inproceedings{redmon2016you,
  title={You only look once: Unified, real-time object detection},
  author={Redmon, Joseph and Divvala, Santosh and Girshick, Ross and Farhadi, Ali},
  booktitle={Proceedings of the IEEE conference on computer vision and pattern recognition},
  pages={779--788},
  year={2016}
}

@article{ren2016faster,
  title={Faster R-CNN: Towards real-time object detection with region proposal networks},
  author={Ren, Shaoqing and He, Kaiming and Girshick, Ross and Sun, Jian},
  journal={IEEE transactions on pattern analysis and machine intelligence},
  volume={39},
  number={6},
  pages={1137--1149},
  year={2016},
  publisher={IEEE}
}

@article{sutskever2014sequence,
  title={Sequence to sequence learning with neural networks},
  author={Sutskever, Ilya and Vinyals, Oriol and Le, Quoc V},
  journal={Advances in neural information processing systems},
  volume={27},
  year={2014}
}

@article{bahdanau2014neural,
  title={Neural machine translation by jointly learning to align and translate},
  author={Bahdanau, Dzmitry and Cho, Kyunghyun and Bengio, Yoshua},
  journal={arXiv preprint arXiv:1409.0473},
  year={2014}
}

@article{brown2020language,
  title={Language models are few-shot learners},
  author={Brown, Tom and Mann, Benjamin and Ryder, Nick and Subbiah, Melanie and Kaplan, Jared D and Dhariwal, Prafulla and Neelakantan, Arvind and Shyam, Pranav and Sastry, Girish and Askell, Amanda and others},
  journal={Advances in neural information processing systems},
  volume={33},
  pages={1877--1901},
  year={2020}
}

@article{ouyang2022training,
  title={Training language models to follow instructions with human feedback},
  author={Ouyang, Long and Wu, Jeffrey and Jiang, Xu and Almeida, Diogo and Wainwright, Carroll and Mishkin, Pamela and Zhang, Chong and Agarwal, Sandhini and Slama, Katarina and Ray, Alex and others},
  journal={Advances in neural information processing systems},
  volume={35},
  pages={27730--27744},
  year={2022}
}

@article{liu2023visual,
  title={Visual instruction tuning},
  author={Liu, Haotian and Li, Chunyuan and Wu, Qingyang and Lee, Yong Jae},
  journal={Advances in neural information processing systems},
  volume={36},
  pages={34892--34916},
  year={2023}
}

@article{chen2021evaluating,
  title={Evaluating large language models trained on code},
  author={Chen, Mark and Tworek, Jerry and Jun, Heewoo and Yuan, Qiming and Pinto, Henrique Ponde De Oliveira and Kaplan, Jared and Edwards, Harri and Burda, Yuri and Joseph, Nicholas and Brockman, Greg and others},
  journal={arXiv preprint arXiv:2107.03374},
  year={2021}
}

@article{yao2022react,
  title={React: Synergizing reasoning and acting in language models},
  author={Yao, Shunyu and Zhao, Jeffrey and Yu, Dian and Du, Nan and Shafran, Izhak and Narasimhan, Karthik and Cao, Yuan},
  journal={arXiv preprint arXiv:2210.03629},
  year={2022}
}

@article{DBLP:journals/jmlr/LevineFDA16,
  author       = {Sergey Levine and
                  Chelsea Finn and
                  Trevor Darrell and
                  Pieter Abbeel},
  title        = {End-to-End Training of Deep Visuomotor Policies},
  journal      = {J. Mach. Learn. Res.},
  volume       = {17},
  pages        = {39:1--39:40},
  year         = {2016},
  url          = {https://jmlr.org/papers/v17/15-522.html},
  bibsource    = {dblp computer science bibliography, https://dblp.org}
}

@inproceedings{DBLP:conf/icra/PintoG16,
  author       = {Lerrel Pinto and
                  Abhinav Gupta},
  editor       = {Danica Kragic and
                  Antonio Bicchi and
                  Alessandro De Luca},
  title        = {Supersizing self-supervision: Learning to grasp from 50K tries and
                  700 robot hours},
  booktitle    = {2016 {IEEE} International Conference on Robotics and Automation, {ICRA}
                  2016, Stockholm, Sweden, May 16-21, 2016},
  pages        = {3406--3413},
  publisher    = {{IEEE}},
  year         = {2016},
  url          = {https://doi.org/10.1109/ICRA.2016.7487517},
  doi          = {10.1109/ICRA.2016.7487517},
  bibsource    = {dblp computer science bibliography, https://dblp.org}
}

@inproceedings{DBLP:conf/rss/BrohanBCCDFGHHH23,
  author       = {Anthony Brohan and
                  Noah Brown and
                  Justice Carbajal and
                  Yevgen Chebotar and
                  Joseph Dabis and
                  Chelsea Finn and
                  Keerthana Gopalakrishnan and
                  Karol Hausman and
                  Alexander Herzog and
                  Jasmine Hsu and
                  Julian Ibarz and
                  Brian Ichter and
                  Alex Irpan and
                  Tomas Jackson and
                  Sally Jesmonth and
                  Nikhil J. Joshi and
                  Ryan Julian and
                  Dmitry Kalashnikov and
                  Yuheng Kuang and
                  Isabel Leal and
                  Kuang{-}Huei Lee and
                  Sergey Levine and
                  Yao Lu and
                  Utsav Malla and
                  Deeksha Manjunath and
                  Igor Mordatch and
                  Ofir Nachum and
                  Carolina Parada and
                  Jodilyn Peralta and
                  Emily Perez and
                  Karl Pertsch and
                  Jornell Quiambao and
                  Kanishka Rao and
                  Michael S. Ryoo and
                  Grecia Salazar and
                  Pannag R. Sanketi and
                  Kevin Sayed and
                  Jaspiar Singh and
                  Sumedh Sontakke and
                  Austin Stone and
                  Clayton Tan and
                  Huong T. Tran and
                  Vincent Vanhoucke and
                  Steve Vega and
                  Quan Vuong and
                  Fei Xia and
                  Ted Xiao and
                  Peng Xu and
                  Sichun Xu and
                  Tianhe Yu and
                  Brianna Zitkovich},
  editor       = {Kostas E. Bekris and
                  Kris Hauser and
                  Sylvia L. Herbert and
                  Jingjin Yu},
  title        = {{RT-1:} Robotics Transformer for Real-World Control at Scale},
  booktitle    = {Robotics: Science and Systems XIX, Daegu, Republic of Korea, July
                  10-14, 2023},
  year         = {2023},
  url          = {https://doi.org/10.15607/RSS.2023.XIX.025},
  doi          = {10.15607/RSS.2023.XIX.025},
  bibsource    = {dblp computer science bibliography, https://dblp.org}
}

@inproceedings{DBLP:conf/rss/GhoshWPBMDHK0LT24,
  author       = {Dibya Ghosh and
                  Homer Rich Walke and
                  Karl Pertsch and
                  Kevin Black and
                  Oier Mees and
                  Sudeep Dasari and
                  Joey Hejna and
                  Tobias Kreiman and
                  Charles Xu and
                  Jianlan Luo and
                  You Liang Tan and
                  Lawrence Yunliang Chen and
                  Quan Vuong and
                  Ted Xiao and
                  Pannag R. Sanketi and
                  Dorsa Sadigh and
                  Chelsea Finn and
                  Sergey Levine},
  editor       = {Dana Kulic and
                  Gentiane Venture and
                  Kostas E. Bekris and
                  Enrique Coronado},
  title        = {Octo: An Open-Source Generalist Robot Policy},
  booktitle    = {Robotics: Science and Systems XX, Delft, The Netherlands, July 15-19,
                  2024},
  year         = {2024},
  url          = {https://doi.org/10.15607/RSS.2024.XX.090},
  doi          = {10.15607/RSS.2024.XX.090},
  bibsource    = {dblp computer science bibliography, https://dblp.org}
}

@inproceedings{DBLP:conf/iclr/LiuWLTCWX0025,
  author       = {Songming Liu and
                  Lingxuan Wu and
                  Bangguo Li and
                  Hengkai Tan and
                  Huayu Chen and
                  Zhengyi Wang and
                  Ke Xu and
                  Hang Su and
                  Jun Zhu},
  title        = {{RDT-1B:} a Diffusion Foundation Model for Bimanual Manipulation},
  booktitle    = {The Thirteenth International Conference on Learning Representations,
                  {ICLR} 2025, Singapore, April 24-28, 2025},
  publisher    = {OpenReview.net},
  year         = {2025},
  url          = {https://openreview.net/forum?id=yAzN4tz7oI},
  bibsource    = {dblp computer science bibliography, https://dblp.org}
}

@InProceedings{pmlr-v229-zitkovich23a,
  title = 	 {RT-2: Vision-Language-Action Models Transfer Web Knowledge to Robotic Control},
  author =       {Zitkovich, Brianna and Yu, Tianhe and Xu, Sichun and Xu, Peng and Xiao, Ted and Xia, Fei and Wu, Jialin and Wohlhart, Paul and Welker, Stefan and Wahid, Ayzaan and Vuong, Quan and Vanhoucke, Vincent and Tran, Huong and Soricut, Radu and Singh, Anikait and Singh, Jaspiar and Sermanet, Pierre and Sanketi, Pannag R. and Salazar, Grecia and Ryoo, Michael S. and Reymann, Krista and Rao, Kanishka and Pertsch, Karl and Mordatch, Igor and Michalewski, Henryk and Lu, Yao and Levine, Sergey and Lee, Lisa and Lee, Tsang-Wei Edward and Leal, Isabel and Kuang, Yuheng and Kalashnikov, Dmitry and Julian, Ryan and Joshi, Nikhil J. and Irpan, Alex and Ichter, Brian and Hsu, Jasmine and Herzog, Alexander and Hausman, Karol and Gopalakrishnan, Keerthana and Fu, Chuyuan and Florence, Pete and Finn, Chelsea and Dubey, Kumar Avinava and Driess, Danny and Ding, Tianli and Choromanski, Krzysztof Marcin and Chen, Xi and Chebotar, Yevgen and Carbajal, Justice and Brown, Noah and Brohan, Anthony and Arenas, Montserrat Gonzalez and Han, Kehang},
  booktitle = 	 {Proceedings of The 7th Conference on Robot Learning},
  pages = 	 {2165--2183},
  year = 	 {2023},
  editor = 	 {Tan, Jie and Toussaint, Marc and Darvish, Kourosh},
  volume = 	 {229},
  series = 	 {Proceedings of Machine Learning Research},
  month = 	 {06--09 Nov},
  publisher =    {PMLR},
  url = 	 {https://proceedings.mlr.press/v229/zitkovich23a.html}
}

@inproceedings{DBLP:conf/corl/KimPKXB0RFSVKBT24,
  author       = {Moo Jin Kim and
                  Karl Pertsch and
                  Siddharth Karamcheti and
                  Ted Xiao and
                  Ashwin Balakrishna and
                  Suraj Nair and
                  Rafael Rafailov and
                  Ethan Paul Foster and
                  Pannag R. Sanketi and
                  Quan Vuong and
                  Thomas Kollar and
                  Benjamin Burchfiel and
                  Russ Tedrake and
                  Dorsa Sadigh and
                  Sergey Levine and
                  Percy Liang and
                  Chelsea Finn},
  editor       = {Pulkit Agrawal and
                  Oliver Kroemer and
                  Wolfram Burgard},
  title        = {OpenVLA: An Open-Source Vision-Language-Action Model},
  booktitle    = {Conference on Robot Learning, 6-9 November 2024, Munich, Germany},
  series       = {Proceedings of Machine Learning Research},
  volume       = {270},
  pages        = {2679--2713},
  publisher    = {{PMLR}},
  year         = {2024},
  url          = {https://proceedings.mlr.press/v270/kim25c.html},
  bibsource    = {dblp computer science bibliography, https://dblp.org}
}

@article{DBLP:journals/corr/abs-2410-24164,
  author       = {Kevin Black and
                  Noah Brown and
                  Danny Driess and
                  Adnan Esmail and
                  Michael Equi and
                  Chelsea Finn and
                  Niccolo Fusai and
                  Lachy Groom and
                  Karol Hausman and
                  Brian Ichter and
                  Szymon Jakubczak and
                  Tim Jones and
                  Liyiming Ke and
                  Sergey Levine and
                  Adrian Li{-}Bell and
                  Mohith Mothukuri and
                  Suraj Nair and
                  Karl Pertsch and
                  Lucy Xiaoyang Shi and
                  James Tanner and
                  Quan Vuong and
                  Anna Walling and
                  Haohuan Wang and
                  Ury Zhilinsky},
  title        = {{\(\pi\)}\({}_{\mbox{0}}\): {A} Vision-Language-Action Flow Model
                  for General Robot Control},
  journal      = {CoRR},
  volume       = {abs/2410.24164},
  year         = {2024},
  url          = {https://doi.org/10.48550/arXiv.2410.24164},
  doi          = {10.48550/ARXIV.2410.24164},
  eprinttype   = {arXiv},
  eprint       = {2410.24164},
  bibsource    = {dblp computer science bibliography, https://dblp.org}
}

@article{DBLP:journals/corr/abs-2504-16054,
  author       = {Physical Intelligence and
                  Kevin Black and
                  Noah Brown and
                  James Darpinian and
                  Karan Dhabalia and
                  Danny Driess and
                  Adnan Esmail and
                  Michael Equi and
                  Chelsea Finn and
                  Niccolo Fusai and
                  Manuel Y. Galliker and
                  Dibya Ghosh and
                  Lachy Groom and
                  Karol Hausman and
                  Brian Ichter and
                  Szymon Jakubczak and
                  Tim Jones and
                  Liyiming Ke and
                  Devin LeBlanc and
                  Sergey Levine and
                  Adrian Li{-}Bell and
                  Mohith Mothukuri and
                  Suraj Nair and
                  Karl Pertsch and
                  Allen Z. Ren and
                  Lucy Xiaoyang Shi and
                  Laura Smith and
                  Jost Tobias Springenberg and
                  Kyle Stachowicz and
                  James Tanner and
                  Quan Vuong and
                  Homer Walke and
                  Anna Walling and
                  Haohuan Wang and
                  Lili Yu and
                  Ury Zhilinsky},
  title        = {{\(\pi\)}\({}_{\mbox{0.5}}\): a Vision-Language-Action Model with
                  Open-World Generalization},
  journal      = {CoRR},
  volume       = {abs/2504.16054},
  year         = {2025},
  url          = {https://doi.org/10.48550/arXiv.2504.16054},
  doi          = {10.48550/ARXIV.2504.16054},
  eprinttype   = {arXiv},
  eprint       = {2504.16054},
  bibsource    = {dblp computer science bibliography, https://dblp.org}
}

@article{DBLP:journals/corr/abs-2503-14734,
  author       = {Johan Bjorck and
                  Fernando Casta{\~{n}}eda and
                  Nikita Cherniadev and
                  Xingye Da and
                  Runyu Ding and
                  Linxi Fan and
                  Yu Fang and
                  Dieter Fox and
                  Fengyuan Hu and
                  Spencer Huang and
                  Joel Jang and
                  Zhenyu Jiang and
                  Jan Kautz and
                  Kaushil Kundalia and
                  Lawrence Lao and
                  Zhiqi Li and
                  Zongyu Lin and
                  Kevin Lin and
                  Guilin Liu and
                  Edith LLontop and
                  Loic Magne and
                  Ajay Mandlekar and
                  Avnish Narayan and
                  Soroush Nasiriany and
                  Scott Reed and
                  You Liang Tan and
                  Guanzhi Wang and
                  Zu Wang and
                  Jing Wang and
                  Qi Wang and
                  Jiannan Xiang and
                  Yuqi Xie and
                  Yinzhen Xu and
                  Zhenjia Xu and
                  Seonghyeon Ye and
                  Zhiding Yu and
                  Ao Zhang and
                  Hao Zhang and
                  Yizhou Zhao and
                  Ruijie Zheng and
                  Yuke Zhu},
  title        = {{GR00T} {N1:} An Open Foundation Model for Generalist Humanoid Robots},
  journal      = {CoRR},
  volume       = {abs/2503.14734},
  year         = {2025},
  url          = {https://doi.org/10.48550/arXiv.2503.14734},
  doi          = {10.48550/ARXIV.2503.14734},
  eprinttype   = {arXiv},
  eprint       = {2503.14734},
  bibsource    = {dblp computer science bibliography, https://dblp.org}
}

@article{DBLP:journals/corr/abs-2503-20020,
  author       = {Gemini Robotics Team},
  title        = {Gemini Robotics: Bringing {AI} into the Physical World},
  journal      = {CoRR},
  volume       = {abs/2503.20020},
  year         = {2025},
  url          = {https://doi.org/10.48550/arXiv.2503.20020},
  doi          = {10.48550/ARXIV.2503.20020},
  eprinttype   = {arXiv},
  eprint       = {2503.20020},
  bibsource    = {dblp computer science bibliography, https://dblp.org}
}

@article{DBLP:journals/corr/abs-2510-10274,
  author       = {Jinliang Zheng and
                  Jianxiong Li and
                  Zhihao Wang and
                  Dongxiu Liu and
                  Xirui Kang and
                  Yuchun Feng and
                  Yinan Zheng and
                  Jiayin Zou and
                  Yilun Chen and
                  Jia Zeng and
                  Ya{-}Qin Zhang and
                  Jiangmiao Pang and
                  Jingjing Liu and
                  Tai Wang and
                  Xianyuan Zhan},
  title        = {{X-VLA:} Soft-Prompted Transformer as Scalable Cross-Embodiment Vision-Language-Action
                  Model},
  journal      = {CoRR},
  volume       = {abs/2510.10274},
  year         = {2025},
  url          = {https://doi.org/10.48550/arXiv.2510.10274},
  doi          = {10.48550/ARXIV.2510.10274},
  eprinttype   = {arXiv},
  eprint       = {2510.10274},
  bibsource    = {dblp computer science bibliography, https://dblp.org}
}

@inproceedings{DBLP:conf/iclr/YangDGTKSA24,
  author       = {Sherry Yang and
                  Yilun Du and
                  Seyed Kamyar Seyed Ghasemipour and
                  Jonathan Tompson and
                  Leslie Pack Kaelbling and
                  Dale Schuurmans and
                  Pieter Abbeel},
  title        = {Learning Interactive Real-World Simulators},
  booktitle    = {The Twelfth International Conference on Learning Representations,
                  {ICLR} 2024, Vienna, Austria, May 7-11, 2024},
  publisher    = {OpenReview.net},
  year         = {2024},
  url          = {https://openreview.net/forum?id=sFyTZEqmUY},
  bibsource    = {dblp computer science bibliography, https://dblp.org}
}

@inproceedings{DBLP:conf/icml/BruceDEPS0LMSAA24,
  author       = {Jake Bruce and
                  Michael D. Dennis and
                  Ashley Edwards and
                  Jack Parker{-}Holder and
                  Yuge Shi and
                  Edward Hughes and
                  Matthew Lai and
                  Aditi Mavalankar and
                  Richie Steigerwald and
                  Chris Apps and
                  Yusuf Aytar and
                  Sarah Bechtle and
                  Feryal M. P. Behbahani and
                  Stephanie C. Y. Chan and
                  Nicolas Heess and
                  Lucy Gonzalez and
                  Simon Osindero and
                  Sherjil Ozair and
                  Scott E. Reed and
                  Jingwei Zhang and
                  Konrad Zolna and
                  Jeff Clune and
                  Nando de Freitas and
                  Satinder Singh and
                  Tim Rockt{\"{a}}schel},
  editor       = {Ruslan Salakhutdinov and
                  Zico Kolter and
                  Katherine A. Heller and
                  Adrian Weller and
                  Nuria Oliver and
                  Jonathan Scarlett and
                  Felix Berkenkamp},
  title        = {Genie: Generative Interactive Environments},
  booktitle    = {Forty-first International Conference on Machine Learning, {ICML} 2024,
                  Vienna, Austria, July 21-27, 2024},
  series       = {Proceedings of Machine Learning Research},
  volume       = {235},
  pages        = {4603--4623},
  publisher    = {{PMLR} / OpenReview.net},
  year         = {2024},
  url          = {https://proceedings.mlr.press/v235/bruce24a.html},
  bibsource    = {dblp computer science bibliography, https://dblp.org}
}

@article{DBLP:journals/corr/abs-2602-15922,
  author       = {Seonghyeon Ye and
                  Yunhao Ge and
                  Kaiyuan Zheng and
                  Shenyuan Gao and
                  Sihyun Yu and
                  George Kurian and
                  Suneel Indupuru and
                  You Liang Tan and
                  Chuning Zhu and
                  Jiannan Xiang and
                  Ayaan Malik and
                  Kyungmin Lee and
                  William Liang and
                  Nadun Ranawaka and
                  Jiasheng Gu and
                  Yinzhen Xu and
                  Guanzhi Wang and
                  Fengyuan Hu and
                  Avnish Narayan and
                  Johan Bjorck and
                  Jing Wang and
                  Gwanghyun Kim and
                  Dantong Niu and
                  Ruijie Zheng and
                  Yuqi Xie and
                  Jimmy Wu and
                  Qi Wang and
                  Ryan Julian and
                  Danfei Xu and
                  Yilun Du and
                  Yevgen Chebotar and
                  Scott Reed and
                  Jan Kautz and
                  Yuke Zhu and
                  Linxi "Jim" Fan and
                  Joel Jang},
  title        = {World Action Models are Zero-shot Policies},
  journal      = {CoRR},
  volume       = {abs/2602.15922},
  year         = {2026},
  url          = {https://doi.org/10.48550/arXiv.2602.15922},
  doi          = {10.48550/ARXIV.2602.15922},
  eprinttype   = {arXiv},
  eprint       = {2602.15922},
  bibsource    = {dblp computer science bibliography, https://dblp.org}
}

@article{DBLP:journals/corr/abs-2604-27792,
  author       = {MotuBrain Team and
                  Chendong Xiang and
                  Fan Bao and
                  Haitian Liu and
                  Hengkai Tan and
                  Hongzhe Bi and
                  James Li and
                  Jiabao Liu and
                  Jingrui Pang and
                  Kiro Jing and
                  Louis Liu and
                  Mengchen Cai and
                  Rongxu Cui and
                  Ruowen Zhao and
                  Runqing Wang and
                  Shuhe Huang and
                  Yao Feng and
                  Yinze Rong and
                  Zeyuan Wang and
                  Jun Zhu},
  title        = {MotuBrain: An Advanced World Action Model for Robot Control},
  journal      = {CoRR},
  volume       = {abs/2604.27792},
  year         = {2026},
  url          = {https://doi.org/10.48550/arXiv.2604.27792},
  doi          = {10.48550/ARXIV.2604.27792},
  eprinttype   = {arXiv},
  eprint       = {2604.27792},
  bibsource    = {dblp computer science bibliography, https://dblp.org}
}

@inproceedings{DBLP:conf/corl/IchterBCFHHHIIJ22,
  author       = {Brian Ichter and
                  Anthony Brohan and
                  Yevgen Chebotar and
                  Chelsea Finn and
                  Karol Hausman and
                  Alexander Herzog and
                  Daniel Ho and
                  Julian Ibarz and
                  Alex Irpan and
                  Eric Jang and
                  Ryan Julian and
                  Dmitry Kalashnikov and
                  Sergey Levine and
                  Yao Lu and
                  Carolina Parada and
                  Kanishka Rao and
                  Pierre Sermanet and
                  Alexander Toshev and
                  Vincent Vanhoucke and
                  Fei Xia and
                  Ted Xiao and
                  Peng Xu and
                  Mengyuan Yan and
                  Noah Brown and
                  Michael Ahn and
                  Omar Cortes and
                  Nicolas Sievers and
                  Clayton Tan and
                  Sichun Xu and
                  Diego Reyes and
                  Jarek Rettinghouse and
                  Jornell Quiambao and
                  Peter Pastor and
                  Linda Luu and
                  Kuang{-}Huei Lee and
                  Yuheng Kuang and
                  Sally Jesmonth and
                  Nikhil J. Joshi and
                  Kyle Jeffrey and
                  Rosario Jauregui Ruano and
                  Jasmine Hsu and
                  Keerthana Gopalakrishnan and
                  Byron David and
                  Andy Zeng and
                  Chuyuan Kelly Fu},
  editor       = {Karen Liu and
                  Dana Kulic and
                  Jeffrey Ichnowski},
  title        = {Do As {I} Can, Not As {I} Say: Grounding Language in Robotic Affordances},
  booktitle    = {Conference on Robot Learning, CoRL 2022, 14-18 December 2022, Auckland,
                  New Zealand},
  series       = {Proceedings of Machine Learning Research},
  volume       = {205},
  pages        = {287--318},
  publisher    = {{PMLR}},
  year         = {2022},
  url          = {https://proceedings.mlr.press/v205/ichter23a.html},
  bibsource    = {dblp computer science bibliography, https://dblp.org}
}

@inproceedings{DBLP:conf/icra/LiangHXXHIFZ23,
  author       = {Jacky Liang and
                  Wenlong Huang and
                  Fei Xia and
                  Peng Xu and
                  Karol Hausman and
                  Brian Ichter and
                  Pete Florence and
                  Andy Zeng},
  title        = {Code as Policies: Language Model Programs for Embodied Control},
  booktitle    = {{IEEE} International Conference on Robotics and Automation, {ICRA}
                  2023, London, UK, May 29 - June 2, 2023},
  pages        = {9493--9500},
  publisher    = {{IEEE}},
  year         = {2023},
  url          = {https://doi.org/10.1109/ICRA48891.2023.10160591},
  doi          = {10.1109/ICRA48891.2023.10160591},
  bibsource    = {dblp computer science bibliography, https://dblp.org}
}

@inproceedings{DBLP:conf/corl/HuangXXCLFZTMCS22,
  author       = {Wenlong Huang and
                  Fei Xia and
                  Ted Xiao and
                  Harris Chan and
                  Jacky Liang and
                  Pete Florence and
                  Andy Zeng and
                  Jonathan Tompson and
                  Igor Mordatch and
                  Yevgen Chebotar and
                  Pierre Sermanet and
                  Tomas Jackson and
                  Noah Brown and
                  Linda Luu and
                  Sergey Levine and
                  Karol Hausman and
                  Brian Ichter},
  editor       = {Karen Liu and
                  Dana Kulic and
                  Jeffrey Ichnowski},
  title        = {Inner Monologue: Embodied Reasoning through Planning with Language
                  Models},
  booktitle    = {Conference on Robot Learning, CoRL 2022, 14-18 December 2022, Auckland,
                  New Zealand},
  series       = {Proceedings of Machine Learning Research},
  volume       = {205},
  pages        = {1769--1782},
  publisher    = {{PMLR}},
  year         = {2022},
  url          = {https://proceedings.mlr.press/v205/huang23c.html},
  bibsource    = {dblp computer science bibliography, https://dblp.org}
}

@inproceedings{DBLP:conf/corl/HuangWLZF24,
  author       = {Wenlong Huang and
                  Chen Wang and
                  Yunzhu Li and
                  Ruohan Zhang and
                  Li Fei{-}Fei},
  editor       = {Pulkit Agrawal and
                  Oliver Kroemer and
                  Wolfram Burgard},
  title        = {ReKep: Spatio-Temporal Reasoning of Relational Keypoint Constraints
                  for Robotic Manipulation},
  booktitle    = {Conference on Robot Learning, 6-9 November 2024, Munich, Germany},
  series       = {Proceedings of Machine Learning Research},
  volume       = {270},
  pages        = {4573--4602},
  publisher    = {{PMLR}},
  year         = {2024},
  url          = {https://proceedings.mlr.press/v270/huang25g.html},
  bibsource    = {dblp computer science bibliography, https://dblp.org}
}

@article{DBLP:journals/corr/abs-2505-23450,
  author       = {Zhejian Yang and
                  Yongchao Chen and
                  Xueyang Zhou and
                  Jiangyue Yan and
                  Dingjie Song and
                  Yinuo Liu and
                  Yuting Li and
                  Yu Zhang and
                  Pan Zhou and
                  Hechang Chen and
                  Lichao Sun},
  title        = {Agentic Robot: {A} Brain-Inspired Framework for Vision-Language-Action
                  Models in Embodied Agents},
  journal      = {CoRR},
  volume       = {abs/2505.23450},
  year         = {2025},
  url          = {https://doi.org/10.48550/arXiv.2505.23450},
  doi          = {10.48550/ARXIV.2505.23450},
  eprinttype   = {arXiv},
  eprint       = {2505.23450},
  bibsource    = {dblp computer science bibliography, https://dblp.org}
}

@article{DBLP:journals/corr/abs-2604-13942,
  author       = {Zhen Liu and
                  Xinyu Ning and
                  Zhe Hu and
                  XinXin Xie and
                  Weize Li and
                  Zhipeng Tang and
                  Chongyu Wang and
                  Zejun Yang and
                  Hanlin Wang and
                  Yitong Liu and
                  Zhongzhu Pu},
  title        = {Goal2Skill: Long-Horizon Manipulation with Adaptive Planning and Reflection},
  journal      = {CoRR},
  volume       = {abs/2604.13942},
  year         = {2026},
  url          = {https://doi.org/10.48550/arXiv.2604.13942},
  doi          = {10.48550/ARXIV.2604.13942},
  eprinttype   = {arXiv},
  eprint       = {2604.13942},
  bibsource    = {dblp computer science bibliography, https://dblp.org}
}

@article{DBLP:journals/corr/abs-2603-22435,
  author       = {Max Fu and
                  Justin Yu and
                  Karim El{-}Refai and
                  Ethan Kou and
                  Haoru Xue and
                  Huang Huang and
                  Wenli Xiao and
                  Guanzhi Wang and
                  Feifei Li and
                  Guanya Shi and
                  Jiajun Wu and
                  Shankar S. Sastry and
                  Yuke Zhu and
                  Ken Goldberg and
                  Linxi "Jim" Fan},
  title        = {CaP-X: {A} Framework for Benchmarking and Improving Coding Agents
                  for Robot Manipulation},
  journal      = {CoRR},
  volume       = {abs/2603.22435},
  year         = {2026},
  url          = {https://doi.org/10.48550/arXiv.2603.22435},
  doi          = {10.48550/ARXIV.2603.22435},
  eprinttype   = {arXiv},
  eprint       = {2603.22435},
  bibsource    = {dblp computer science bibliography, https://dblp.org}
}

@article{DBLP:journals/corr/abs-2607-08448,
  author       = {Yixian Zhang and
                  Huanming Zhang and
                  Feng Gao and
                  Xiao Li and
                  Zhihao Liu and
                  Chunyang Zhu and
                  Jiaxing Qiu and
                  Yuchen Yan and
                  Jiyuan Liu and
                  Wenhao Tang and
                  Zhengru Fang and
                  Yi Nie and
                  Changxu Wei and
                  Yu Wang and
                  Wenbo Ding and
                  Chao Yu},
  title        = {Harness {VLA:} Steering Frozen VLAs into Reliable Manipulation Primitives
                  via Memory-Guided Agents},
  journal      = {CoRR},
  volume       = {abs/2607.08448},
  year         = {2026},
  url          = {https://doi.org/10.48550/arXiv.2607.08448},
  doi          = {10.48550/ARXIV.2607.08448},
  eprinttype   = {arXiv},
  eprint       = {2607.08448},
  bibsource    = {dblp computer science bibliography, https://dblp.org}
}

@misc{chen2026showharnessjustvlmagent,
  title         = {Show-Harness: Just a VLM Agent Can Play Robots},
  author        = {Yanzhe Chen and Zechen Bai and Zhijun Cao and Wenzheng Zeng and Kevin Qinghong Lin and Yiqi Lin and Guoqiang Liang and Kevin Yuchen Ma and Qiming Huang and Mike Zheng Shou},
  year          = {2026},
  eprint        = {2609.10522},
  archivePrefix = {arXiv},
  primaryClass  = {cs.RO},
  url           = {https://arxiv.org/abs/2609.10522}
}

@article{chen2025robotwin,
  title={Robotwin 2.0: A scalable data generator and benchmark with strong domain randomization for robust bimanual robotic manipulation},
  author={Chen, Tianxing and Chen, Zanxin and Chen, Baijun and Cai, Zijian and Liu, Yibin and Li, Zixuan and Liang, Qiwei and Lin, Xianliang and Ge, Yiheng and Gu, Zhenyu and others},
  journal={arXiv preprint arXiv:2506.18088},
  year={2025}
}

@article{chen2026robodojo,
  title={RoboDojo: A Unified Sim-and-Real Benchmark for Comprehensive Evaluation of Generalist Robot Manipulation Policies},
  author={Chen, Tianxing and Chen, Yue and Li, Zixuan and Tang, Junyuan and Su, Kailun and Lu, Haoran and Wan, Weijie and Chen, Baijun and Liu, Songling and Yan, Haowen and others},
  journal={arXiv preprint arXiv:2607.04434},
  year={2026}
}

@article{cai2026xiaomi,
  title={Xiaomi-robotics-0: An open-sourced vision-language-action model with real-time execution},
  author={Cai, Rui and Guo, Jun and He, Xinze and Jin, Piaopiao and Li, Jie and Lin, Bingxuan and Liu, Futeng and Liu, Wei and Ma, Fei and Ma, Kun and others},
  journal={arXiv preprint arXiv:2602.12684},
  year={2026}
}

@article{guo2026unified,
  title={Unified 4d world action modeling from video priors with asynchronous denoising},
  author={Guo, Jun and Li, Qiwei and Li, Peiyan and Chen, Zilong and Sun, Nan and Su, Yifei and Wang, Heyun and Zhang, Yuan and Li, Xinghang and Liu, Huaping},
  journal={arXiv preprint arXiv:2604.26694},
  year={2026}
}

@article{yuan2026fast,
  title={Fast-wam: Do world action models need test-time future imagination?},
  author={Yuan, Tianyuan and Dong, Zibin and Liu, Yicheng and Zhao, Hang},
  journal={arXiv preprint arXiv:2603.16666},
  year={2026}
}

@article{jiang2025galaxea,
  title={Galaxea open-world dataset and g0 dual-system vla model},
  author={Jiang, Tao and Yuan, Tianyuan and Liu, Yicheng and Lu, Chenhao and Cui, Jianning and Liu, Xiao and Cheng, Shuiqi and Gao, Jiyang and Xu, Huazhe and Zhao, Hang},
  journal={arXiv preprint arXiv:2509.00576},
  year={2025}
}

@inproceedings{hancock2026actions,
  title={Actions as language: Fine-tuning vlms into vlas without catastrophic forgetting},
  author={Hancock, Asher and Wu, Xindi and Zha, Lihan and Russakovsky, Olga and Majumdar, Anirudha},
  booktitle={International Conference on Learning Representations},
  volume={2026},
  pages={75333--75360},
  year={2026}
}

@inproceedings{yang2026vision,
  title={Vision-language-action instruction tuning: From understanding to manipulation},
  author={Yang, Shuai and Li, Hao and Wang, Bin and Chen, Yilun and Tian, Yang and Wang, Tai and Wang, Hanqing and Zhao, Feng and Liao, Yiyi and Pang, Jiangmiao},
  booktitle={International Conference on Learning Representations},
  volume={2026},
  pages={152896--152943},
  year={2026}
}

@article{yu2022using,
  title={Using both demonstrations and language instructions to efficiently learn robotic tasks},
  author={Yu, Albert and Mooney, Raymond J},
  journal={arXiv preprint arXiv:2210.04476},
  year={2022}
}

@article{wu2026pragmatic,
  title={A pragmatic vla foundation model},
  author={Wu, Wei and Lu, Fan and Wang, Yunnan and Yang, Shuai and Liu, Shi and Wang, Fangjing and Zhu, Qian and Sun, He and Wang, Yong and Ma, Shuailei and others},
  journal={arXiv preprint arXiv:2601.18692},
  year={2026}
}

@article{community2026starvla,
  title={StarVLA: A Lego-like Codebase for Vision-Language-Action Model Developing},
  author={Community, StarVLA},
  journal={arXiv preprint arXiv:2604.05014},
  year={2026}
}

@article{zhang2026hy,
  title={Hy-embodied-0.5-vla: From vision-language-action models to a real-world robot learning stack},
  author={Zhang, He and Xiang, Lingzhu and Lin, Haitao and Huang, Zeyu and Wang, Minghui and Zhong, Dingyan and Dong, Yubo and Wu, Yihao and Rao, Yongming and Zhang, Dongsheng and others},
  journal={arXiv preprint arXiv:2606.14409},
  year={2026}
}

@article{liu2026g0,
  title={G0. 5: One Autoregressive Stream for Robot Reasoning and Action},
  author={Liu, Yicheng and Dong, Zibin and Ye, Baijun and Yuan, Tianyuan and Jiang, Tao and Yang, Anqi and Cao, Shicheng and Liu, Haonan and Sun, Yue and Guo, Zihan and others},
  journal={arXiv preprint arXiv:2608.11739},
  year={2026}
}

@article{dm05,
  title={DM0.5: Designed for the Open World, Where Generalization Emerges},
  author={Dexmal},
  journal={https://www.dexmal.com/blog/dm0.5},
  year={2026}
}

@misc{xiaomir1,
      title={Xiaomi-Robotics-1: Scaling Vision-Language-Action Models with over 100K Hours of Real-World Trajectories}, 
      author={Xiaomi Robotics Team and Jun Guo and Piaopiao Jin and Jason Li and Peiyan Li and Yingyan Li and Futeng Liu and Wanli Peng and Optimus Qin and Yifei Su and Nan Sun and Qiao Sun and Runze Suo and Heyun Wang and Yunhong Wang and Rujie Wu and Caoyu Xia and Lina Zhang and Jack Zhao and Guoliang Chen and Wenlong Chen and Xinze He and Bin Li and Qing Li and Zhuorong Li and Heng Qu and Wenxuan Song and Diyun Xiang and Yifan Xie and Peiran Xu and Hangjun Ye and Wen Ye and Han Zhao and Quanyun Zhou},
      year={2026},
      eprint={2607.15330},
      archivePrefix={arXiv},
      primaryClass={cs.RO},
      url={https://arxiv.org/abs/2607.15330}, 
}

@inproceedings{ge2025v2pe,
  title={V2pe: Improving multimodal long-context capability of vision-language models with variable visual position encoding},
  author={Ge, Junqi and Chen, Ziyi and Lin, Jintao and Zhu, Jinguo and Liu, Xihui and Dai, Jifeng and Zhu, Xizhou},
  booktitle={2025 IEEE/CVF International Conference on Computer Vision (ICCV)},
  pages={21070--21084},
  year={2025},
  organization={IEEE}
}

@misc{zhang2026gpt6astrarobodojo,
  title  = {An Unexpected Robot Policy: Early Evaluations of GPT-6 Astra on RoboDojo and Beyond},
  author = {Zhang, Wenbo and Wang, Kaixuan and Ouyang, Yutao and
            Huang, Xiaoyu and Li, Liyang and Su, Kailun and
            Jin, Weiyang and Chai, Wenhao and Liang, Haotian and
            Dou, Zhiyang and Chen, Yue and Chen, Tianxing},
  year   = {2026},
  url    = {https://robodojo-benchmark.com/gpt6-astra-robodojo},
  note   = {Equal contribution: Wenbo Zhang, Kaixuan Wang, Yutao Ouyang.
            Project lead: Wenbo Zhang.
            Corresponding: Yue Chen, Tianxing Chen}
}
